\documentclass[10pt,twocolumn,letterpaper]{article}
\usepackage[margin=0.75in,top=1in]{geometry}
\usepackage{placeins}

\usepackage{array}
\usepackage[caption=false,font=normalsize,labelfont=sf,textfont=sf]{subfig}
\usepackage{textcomp}
\usepackage{stfloats}
\usepackage{url}
\usepackage{verbatim}
\usepackage{graphicx}
\usepackage{cite}
\usepackage{amsmath,amssymb,amsfonts,mathtools}
\usepackage{textcomp}
\usepackage{booktabs}
\usepackage{cite}
\usepackage{textcomp}
\usepackage{tablefootnote}
\usepackage[dvipsnames]{xcolor}
\usepackage{cuted}
\usepackage{capt-of}
\usepackage{graphicx,verbatim}
\usepackage{lineno,hyperref}
\usepackage{multicol, multirow}
\usepackage[square,sort,comma,numbers]{natbib}
\usepackage{diagbox}
\usepackage{pifont}
\usepackage{fancyhdr,graphicx}
\usepackage{bm}
\begin{document}

\title{Particle Diffusion Matching: Random Walk Correspondence Search for the Alignment of Standard and Ultra-Widefield Fundus Images}

\author{
Kanggeon Lee$^{1}$, Soochahn Lee$^{2}$\thanks{Corresponding authors}, Kyoung Mu Lee$^{1}$\footnotemark[1] \\
$^{1}$ASRI, Dept. of ECE, Seoul National University, Korea \\
$^{2}$School of Electrical Engineering, Kookmin University, Korea \\
\footnotesize\texttt{dlrkdrjs97@snu.ac.kr, sclee@kookmin.ac.kr, kyoungmu@snu.ac.kr}
}

\date{}
\maketitle

\begin{abstract}
We propose a robust alignment technique for Standard Fundus Images (SFIs) and Ultra-Widefield Fundus Images (UWFIs), which are challenging to align due to differences in scale, appearance, and the scarcity of distinctive features. Our method, termed Particle Diffusion Matching (PDM), performs alignment through an iterative Random Walk Correspondence Search (RWCS) guided by a diffusion model. At each iteration, the model estimates displacement vectors for particle points by considering local appearance, the structural distribution of particles, and an estimated global transformation, enabling progressive refinement of correspondences even under difficult conditions. PDM achieves state-of-the-art performance across multiple retinal image alignment benchmarks, showing substantial improvement on a primary dataset of SFI-UWFI pairs and demonstrating its effectiveness in real-world clinical scenarios. By providing accurate and scalable correspondence estimation, PDM overcomes the limitations of existing methods and facilitates the integration of complementary retinal image modalities. This diffusion-guided search strategy offers a new direction for improving downstream supervised learning, disease diagnosis, and multi-modal image analysis in ophthalmology.
\end{abstract}

\noindent\textbf{Keywords:} Iterative Random Walk, Diffusion Model, Retinal Image Alignment.

\begin{figure*}[!htbp]
    \centering
    \includegraphics[width=1.0\textwidth]{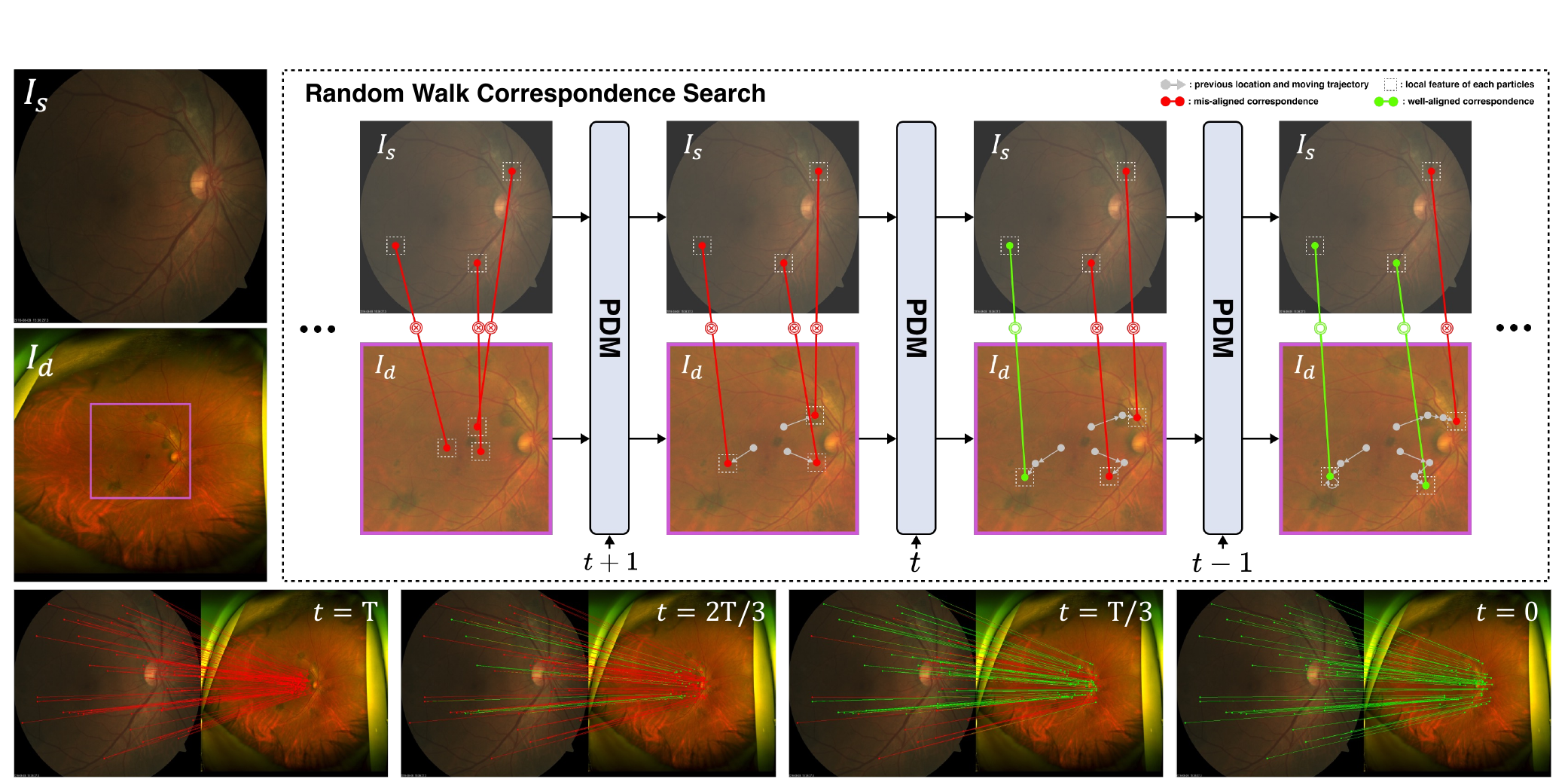}
    \caption{
    \textbf{Random Walk Correspondence Search (RWCS).} 
    Particles are defined as pairs of points from SFI and UWFI, respectively. 
    Particle random walks at sample steps \(t+1\), \(t\), and \(t-1\) are depicted.
    Particle with correct and incorrect correspondences are marked by green and red lines, respectively. 
    The bottom row depicts particles at one-third intervals of the entire sampling process.
    } 
    \label{fig:FIG_SAMP}
\end{figure*}

\section{Introduction}
\begin{figure}[!htbp]
    \centering
    \includegraphics[width=1.0\columnwidth]{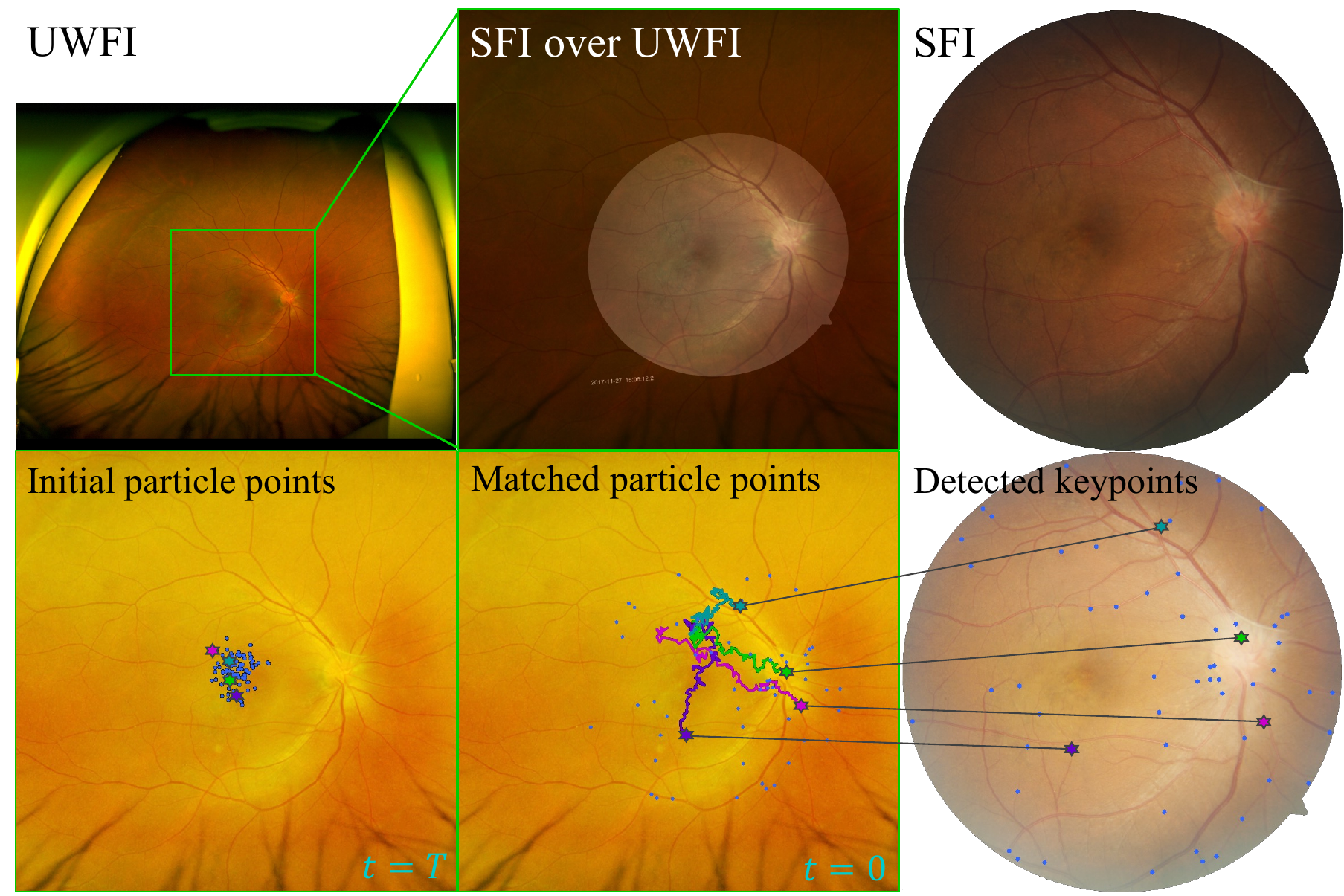}
    \caption{
        \textbf{\emph{Particle Diffusion Matching} (\textsc{PDM}).}
        \textsc{PDM} can align Standard Fundus Images (SFIs) and Ultra-Widefield Fundus Images (UWFIs), which may have extreme differences in view and scale
        (The resolutions of UWFI and SFI here are approximately $2600 \times 2000$ and $3600 \times 3600$, respectively.).
        In \textsc{PDM}, a particle is defined as a point pair comprising a keypoint from the SFI and a potential matching point from the UWFI.
        UWFI points are initialized as random Gaussian points at $t=T$, (bottom left) and undergo random walk correspondence search (RWCS) to eventually reach the matching points (bottom middle) to the paired SFI points (bottom right).
        The random walk trajectories of four sample particles have been highlighted in various colors.
        }
    \label{fig:FIG_TEASER}
\end{figure}

Standard Fundus Images (SFIs) provide a central view of only 30$^\circ$ to 60$^\circ$, covering less than 20\% of the retina~\cite{lee2016ultra}. 
In contrast, Ultra-Widefield Fundus Images (UWFIs) capture up to 200$^\circ$ or 82\% of the retina in a single image~\cite{lee2016ultra, witmer2013comparison}. 
This expanded view, as illustrated in Fig.~\ref{fig:FIG_TEASER}, is crucial for detecting peripheral retinal conditions such as diabetic retinopathy and retinal vascular occlusions.

Despite their broader coverage, UWFI often compromises resolution and clarity compared to SFI, limiting its effectiveness in diagnosing conditions such as age-related macular degeneration and diabetic retinopathy. 
Enhancing UWFI through machine learning techniques, such as image enhancement~\cite{lee2023deep} and super-resolution~\cite{lim2017enhanced}, could potentially allow them to replace SFI. 
However, this requires a training dataset of precisely aligned SFI and UWFI pairs.

The significant differences in field of view and scale, along with variations in color and the lack of distinctive retinal textures, make aligning SFI and UWFI particularly challenging. 
Point correspondence search methods~\cite{lowe2004distinctive,bay2006surf,detone2018superpoint,truong2019glampoints,sarlin2020superglue,liu2022semi} or global transformation estimation methods~\cite{zhao2021deep,cao2022iterative,liu2023geometrized,zhu2024mcnet} struggle when distinctive features are sparse.
Conversely, local deformation estimation methods~\cite{balakrishnan2019voxelmorph,kim2021cyclemorph,kim2022diffusemorph,xu2022gmflow,huang2022flowformer} often struggle to handle the large-scale global transformations typically present in SFI and UWFI alignment tasks.
Although combining global and local estimations in a two-step cascade can improve performance~\cite{de2019deep, lee2019istn}, such methods lack mechanisms to refine or correct errors in the initial global transformation estimates.
Existing methods for retinal images~\cite{liu2022semi,liu2023geometrized,sivaraman2024retinaregnetzeroshotapproachretinal} still struggle with large perspective shifts and local deformations.

We introduce a novel image alignment method designed specifically for image pairs that exhibit significant global transformations alongside subtle local deformations, such as SFI and UWFI pairs. 
Recent work has demonstrated diffusion models' effectiveness in correspondence problems where PoseDiffusion~\cite{wang2023posediffusion} showed that iterative refinement naturally handles pose distributions while capturing inter-point dependencies, and Diff-Reg~\cite{wu2024diff} leveraged diffusion's smooth gradients and initialization robustness for registration tasks.
Inspired by these successes, we develop a correspondence search method where keypoint detection is performed only on the source image, and their corresponding points in the destination image are discovered through an iterative diffusion-guided random walk process.
By incorporating a reverse diffusion process~\cite{ho2020denoising, chan2024tutorial} into this random walk, our method learns to effectively search for accurate correspondence points through continuous refinement rather than discrete matching decisions.
We refer to our approach as \emph{Particle Diffusion Matching} (\textsc{PDM}) to emphasize its distinctive methodology. 
In this framework, keypoints in the destination image, termed \emph{particles} to distinguish them from source image keypoints, undergo a diffusion process that progressively refines their positions during correspondence search.

The key advantage of \textsc{PDM} is its ability to jointly search for corresponding keypoints, which means both maintaining source and destination point pairs as coupled entities and optimizing them collectively.
For images with limited textures or different FOV, conventional methods may yield a high proportion of outliers, for which further processes such as RANSAC~\cite{fischler1981random} may be ineffective. 
\textsc{PDM} does not perform tentative matching, but jointly identifies the set of corresponding points through an iterative diffusion walk.
Unlike conventional methods that treat keypoints independently, each particle in \textsc{PDM} represents a complete correspondence hypothesis from initialization, maintaining an explicit source-to-destination pairing throughout the search process.
During the diffusion walk, these paired particles evolve simultaneously, where each update considers both the individual correspondence's quality and the global geometric coherence among all correspondence pairs.
This joint optimization enables \textsc{PDM} to learn to search for corresponding point sets that produce a geometrically consistent transform function while ensuring more robust 1:1 matching.
By reducing dependence on tentative matches for initial global estimations, \textsc{PDM} effectively minimizes the adverse effects of outliers.
Experimental results confirm the effectiveness of our method, demonstrating its ability to produce more robust and accurate correspondence point estimates.
This leads to enhanced registration accuracy, particularly in challenging scenarios including retinal image alignment, where features are sparse and conventional techniques often fail.

We propose \textsc{PDM} as a novel method for achieving accurate automatic alignment between SFI and UWFI, addressing limitations of previous approaches that relied on unpaired domain adaptation~\cite{ju2020bridge, yoo2020deep} or manual alignment~\cite{thuma2023big}.
Quantitative experimental evaluations demonstrate the superior performance of \textsc{PDM} for retinal image alignment across three datasets: the primary dataset of SFI-UWFI pairs, together with both SFI-SFI and UWFI-UWFI pairs. 
Our method achieves remarkable improvements on the challenging KBSMC dataset with 58.56\% Acceptable rate, representing a 17.65\% absolute improvement over the existing state-of-the-art method~\cite{ren2025minima} (40.91\%) and more than double the performance of traditional correspondence method~\cite{sun2021loftr} (26.20\%). 
Even against recent state-of-the-art approaches specifically designed for cross-modal matching~\cite{ren2025minima} or retinal registration~\cite{sivaraman2024retinaregnetzeroshotapproachretinal}, \textsc{PDM} maintains consistent superiority, validating that our particle diffusion framework fundamentally advances the state of correspondence-based image alignment.

Overall, our contributions are as follows:
\begin{itemize}
    \item We introduce \textsc{PDM}, a novel correspondence framework using diffusion-guided particle optimization with joint geometric constraints.
    \item We achieve state-of-the-art performance on challenging cross-modal retinal image registration tasks.
    \item We demonstrate generalizability through comprehensive evaluation across diverse datasets and thorough ablation studies.
\end{itemize}

\begin{figure}[!htbp] 
    \centering
    \includegraphics[width=1.0\columnwidth]{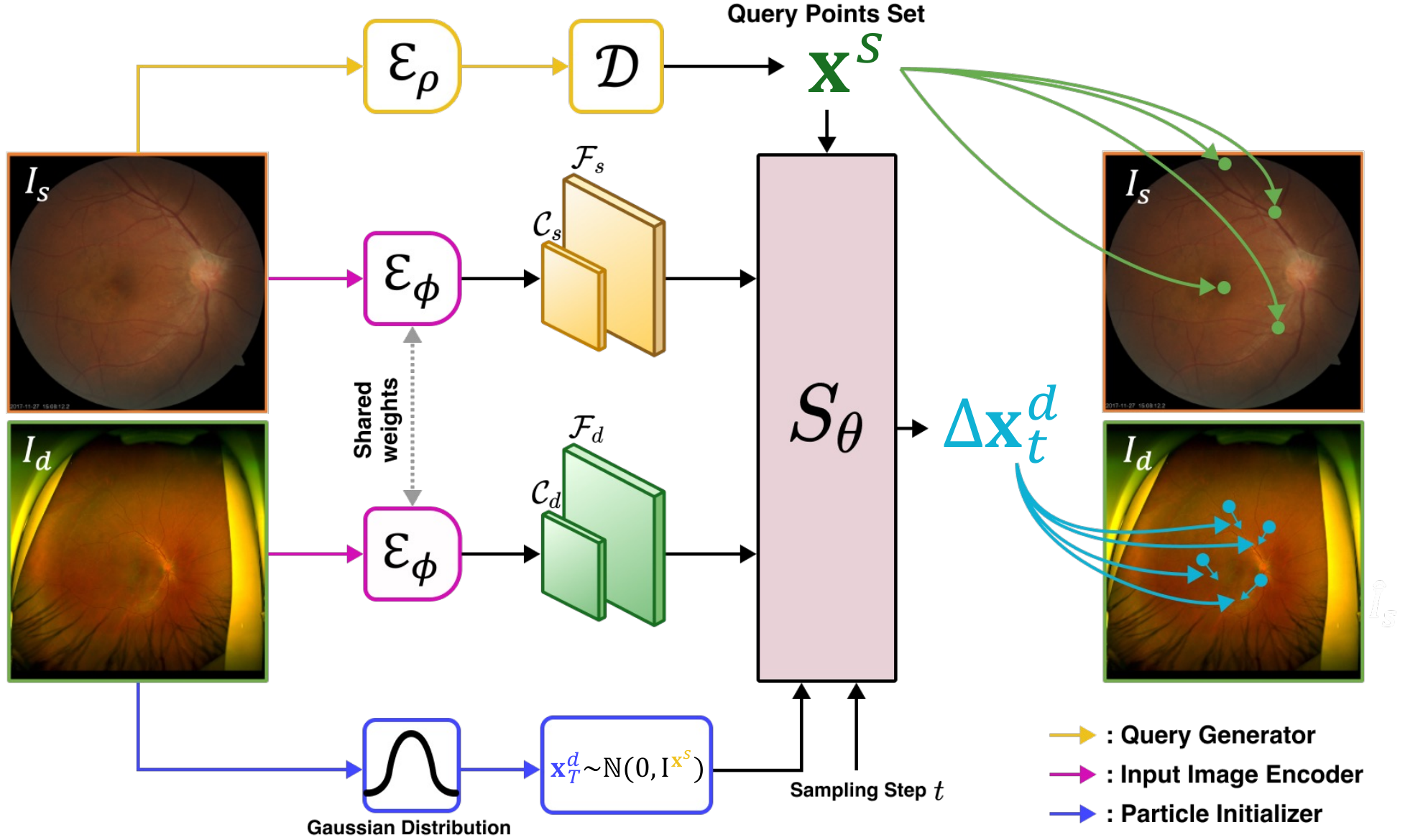}
    \caption{
    \textbf{The architecture of \textsc{PDM}.}
    Given a source image \(I_s\) from SFI and a destination image \(I_d\) from UWFI, the random walker \(\mathbf{S}_\theta\) iteratively refines the correspondence point set \( \mathbf{x}_t^d\) by computing the displacement \(\Delta \mathbf{x}_t^d\) corresponding to the query point set \( \mathbf{x}^s\), which is generated by the key point detector \(\mathcal{D}\). 
    This process starts from initial random Gaussian points \( \mathbf{x}_T^d\) and progressively improves the alignment.
    }
    \label{fig:FIG_NETWORK}
\end{figure}

\section{Related Works}
\label{sec:related}
In this section, we review prior studies relevant to our work, covering global transformation and local deformation estimation, retinal image alignment, and recent advances in alignment methods based on diffusion models.

\noindent
\textbf{Global and Local Transformation Estimation.}  
Traditional global transformation estimation, such as homography computation has relied heavily on handcrafted keypoint detectors and descriptors~\cite{lowe2004distinctive,bay2006surf,rosten2008faster,calonder2010brief}, which often perform poorly under weak local features or extreme appearance variations. 
Learning-based detectors and matchers have enhanced robustness~\cite{detone2018superpoint,truong2019glampoints,10643351,revaud2019r2d2,sarlin2020superglue,lindenberger2023lightglue}, while detector-free approaches directly predict dense correspondences~\cite{detone2016deep,rocco2020ncnet,sun2021loftr,jiang2021cotr,zhang2022relpose,sinha2023sparsepose,tuzcuouglu2024xoftr,edstedt2024roma,ren2025minima,li2025comatchdynamiccovisibilityawaretransformer}. 
However, these methods can struggle with large geometric distortions or domain shifts. 
Local deformation estimation methods primarily address spatially-varying transformations, typically based on optical flow~\cite{dosovitskiy2015flownet,zhang2021separable,xu2022gmflow,huang2022flowformer} or through supervised and unsupervised models developed for medical imaging~\cite{cao2017deformable,balakrishnan2019voxelmorph,hu2018weakly,xu2019deepatlas,kim2021cyclemorph,10158729}. 
Coarse-to-fine strategies~\cite{weinzaepfel2013deepflow} enhance robustness to large deformations, but generally assume that inputs are roughly aligned, leading to suboptimal performance in the presence of significant global misalignments. 
Previous approaches that combine global and local alignment~\cite{jaderberg2015spatial,lee2019istn} or adapt deep adaptive registration~\cite{de2019deep} primarily rely on sequential and independently operated  modules.
In contrast to prior methods, \textsc{PDM} performs joint global-local alignment through a stochastic, iterative search process that progressively refines transformations. 
It uniquely integrates the estimation of both global misalignment and fine-grained local deformations within a unified framework, drawing inspiration from classical iterative strategies~\cite{fischler1981random,besl1992method,cootes1995active} and recent refinement-based approaches~\cite{dong2018learning,zhao2021deep,cao2022iterative,cao2023recurrent,zhu2024mcnet,deng2024crosshomo}.

\noindent
\textbf{Retinal Image Alignment.}  
Retinal image registration has predominantly focused on intra-modality alignment, such as SFI-SFI pairs, leveraging shape models or learning-based keypoint matching methods followed by local refinement~\cite{hernandez2020rempe,liu2022semi,liu2023geometrized,10821896}. 
The recently proposed RetinaRegNet~\cite{sivaraman2024retinaregnetzeroshotapproachretinal} employs diffusion model priors for zero-shot alignment but encounters difficulties when aligning image pairs with substantial style differences. 
Cross-modal registration methods involving SFI and modalities such as OCT or FA~\cite{lee2019adeep,noh2019fine} typically rely on strong supervision or manual preprocessing. 
Similarly, existing approaches for SFI-UWFI alignment~\cite{bioengineering11060568} depend on manual landmark selection. 
To the best of our knowledge, our method is the first fully automatic approach for SFI-UWFI alignment that jointly estimates both global and local transformations.

\noindent
\textbf{Diffusion Models for Alignment.}  
Diffusion models, originally developed for image synthesis~\cite{ho2020denoising,dhariwal2021diffusion,rombach2022high}, have recently been adapted for estimation tasks, including local deformation~\cite{kim2022diffusemorph}, camera pose estimation~\cite{wang2023posediffusion,zhang2024raydiffusion}, modal matching~\cite{wu2024diff}, and structure-from-motion~\cite{zhao2025diffusionsfm}. 
Similar to DiffusionSFM~\cite{zhao2025diffusionsfm}, which unifies 3D camera pose estimation and 3D reconstruction within a single diffusion framework, \textsc{PDM} jointly models global alignment and local deformation through a unified process of progressive refinement, effectively mitigating the limitations commonly associated with conventional two-stage pipelines.

\begin{figure}[!htbp] 
    \centering
    \includegraphics[width=1.0\columnwidth]{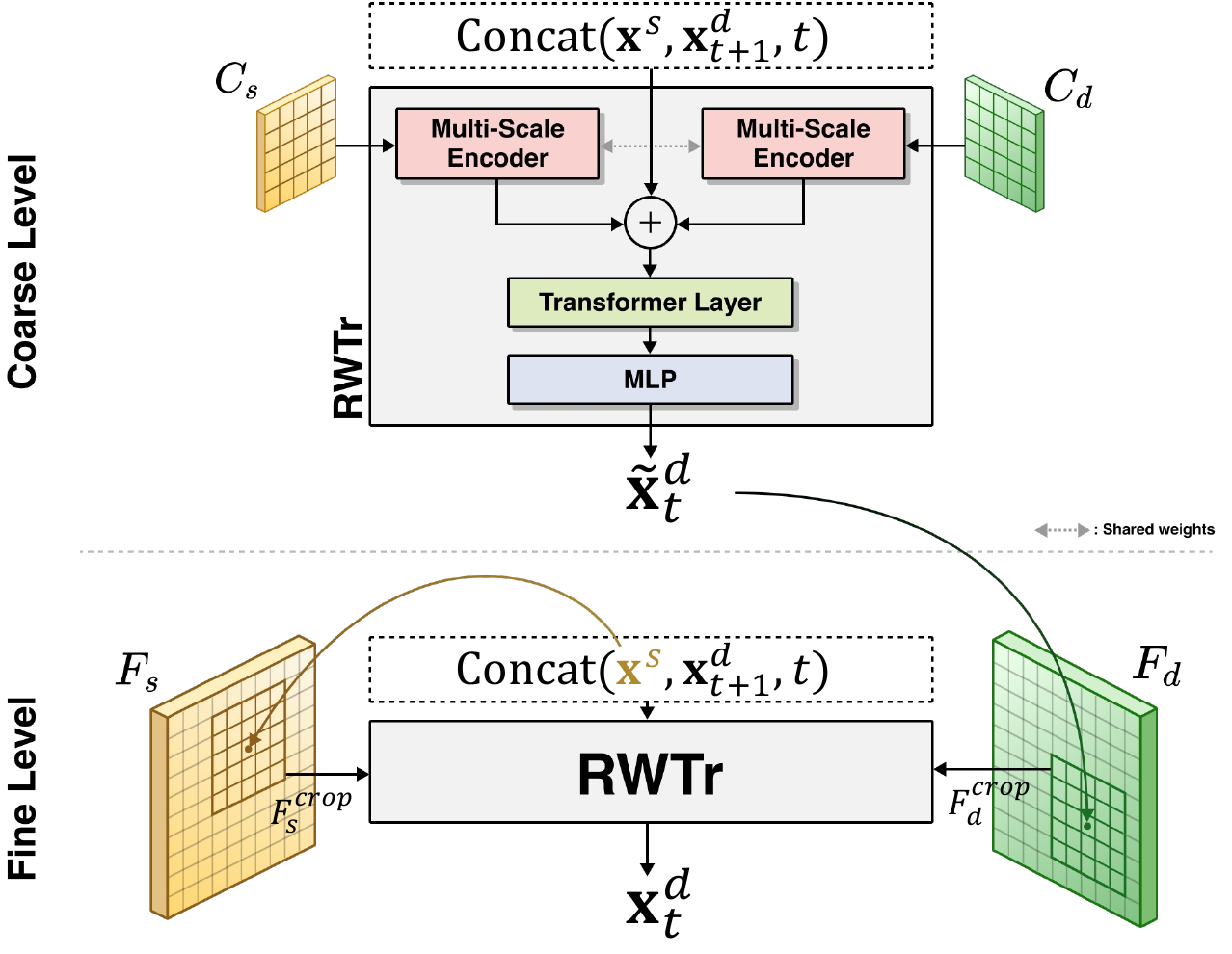}
    \caption{
    \textbf{Detailed structure of \(\mathbf{S}_\theta\).}
    Two \texttt{RWTr}s with the same structure form \(\mathbf{S}_\theta\), where the estimated displacement of the coarse level corresponding points \(\mathbf{\tilde{x}}_t^d\) for the coarse-level features \(C_s\) and \(C_d\) contributes to the local feature selection of the fine-level features \(F_s^{crop}\) and \(F_d^{crop}\) extracted from \(F_s\) and \(F_d\), ultimately estimating the displacement of the fine level corresponding points \(\mathbf{x}_t^d\).
    }
    \label{fig:FIG_RWTR}
\end{figure}

\section{Proposed Method}\label{sec:method}
Here, we provide a detailed explanation of \textsc{PDM} for the random walk correspondence search.

\subsection{Random Walk Correspondence Search}\label{sec:RWCS}
The given source SFI and destination UWFI are denoted as \(I_s\) and \(I_d\). 
We define a particle as a pair of points $(x^{s}_{i}, x^{d}_{i,t})$, with each point, respectively, corresponding to \(I_s\) and \(I_d\), $i$ denoting the particle index, and $t$ denoting the timestep of the random walk diffusion.
We denote the sets of particle points from the \(I_s\) and \(I_d\) portions as ${\mathbf{x}^{s}}=\left\{x^{s}_{i} ~\vert ~i = 0, 1, \dots , N-1 \right\}$ and ${\mathbf{x}^{d}_{t}}=\left\{x^{d}_{i,t} ~\vert ~i = 0, 1, \dots , N-1 \right\}$, respectively, with $N$ denoting the number of particles.

At $t=0$, $\mathbf{x}^{s}$ is defined as the query points detected in \(I_s\) and $\mathbf{x}^{d}_{0}$ are the corresponding points in \(I_d\).
During forward diffusion, points $\mathbf{x}^{d}_{t}$ are diffused, resulting in a Gaussian distribution at $t=T$.
We term the reverse diffusion process as the \emph{random walk correspondence search} (RWCS), as random Gaussian points $\mathbf{x}^{d}_{T}$ are walked until they find positions corresponding to their counterpart query points $\mathbf{x}^{s}$.
The RWCS is powered by a trained diffusion neural network model $\mathbf{S}_{\theta^{***}}$ with parameters $\theta^{***}$, which estimate the motion vectors $\Delta\mathbf{x}^{d}_{t} = \mathbf{S}_{\theta^{***}}(\mathbf{x}^{d}_{t},t \vert I_s,I_d,\mathbf{x}^{s})$ for $\mathbf{x}^{d}_{t}$.
During the RWCS, $\mathbf{x}^{d}_{t}$ are iteratively updated as:
\begin{equation}
   \mathbf{x}^{d}_{t-1} = a\mathbf{x}^{d}_{t} + b\mathbf{S}_{\theta^{***}}(\mathbf{x}^{d}_{t}, t \vert I_{s}, I_{d}, \mathbf{x}^{s}) + c\mathbf{z}
    \label{eq:RWCS_update}
\end{equation}
with coefficients $a,b,c$, and $\mathbf{z}$ denoting random noise of the RWCS. 
A visual representation of RWCS is shown in Fig.~\ref{fig:FIG_SAMP}.

\subsection{The Particle Diffusion Model}\label{sec:score}

Diffusion models~\cite{sohldickstein2015deep,ho2020denoising,song2021scorebased,dhariwal2021diffusionmodelsbeatgans} are latent variable models,
for modeling complex probability distributions and generating corresponding samples.
The diffusion process of a denoising diffusion probabilistic model (DDPM)~\cite{ho2020denoising} is a Markov chain that adds noise to the input data \(\mathbf{x}^{d}_{0}\) defined as follows: 
\begin{equation}\label{eq_ddpm_q1}
\begin{split}
q(\mathbf{x}^{d}_{t}|\mathbf{x}^{d}_{t-1}) & =\mathcal{N}(\mathbf{x}^{d}_{t};\sqrt{1-\beta_{t}}\mathbf{x}^{d}_{t-1},\beta_{t}\mathbf{I}),
\end{split}
\end{equation}
where 
\(\mathcal{N}\) is a Gaussian distribution made to be isotropic by variance schedule \(\beta_{1},...,\beta_{T}\). 
Through a reparameterization trick, \(\mathbf{x}^{d}_{t}\) can directly be sampled from data input \(\mathbf{x}^{d}_{0}\) as:
\begin{equation}\label{eq_ddpm_q2}
\begin{split}
\mathbf{x}^{d}_{t} =\sqrt{\bar{\alpha}_{t}}\mathbf{x}^{d}_{0}+\sqrt{1-\bar{\alpha}_{t}}\mathbf{\epsilon},
\end{split}
\end{equation}
where \(\alpha_{t}=1-\beta_{t}\), \(\bar{\alpha}_{t}=\prod_{t=1}^{T}\alpha_{t}\), and \(\mathbf{\epsilon} \sim \mathcal{N}(0, \mathbf{I})\).

These samples are used to train the neural network to approximate the conditioned probability distributions in the reverse diffusion process:
\begin{equation}\label{eq_ddpm_p}
\begin{split}
p_\theta(\mathbf{x}^{d}_{t-1} \vert \mathbf{x}^{d}_t) & = \mathcal{N}(\mathbf{x}^{d}_{t-1}; \mu_{t}, \Sigma_{t}).
\end{split}
\end{equation}
with the network $\mathbf{S}_{\theta}(\mathbf{x}^{d}_{t}, t \vert I_{s}, I_{d}, \mathbf{x}^{s})$ trained to estimate $\epsilon$, and thus compute the estimate $\boldsymbol{\mu}$ as:
\begin{equation}
\begin{split}
\boldsymbol{\mu}_\theta(\mathbf{x}_t, t) &= {\frac{1}{\sqrt{\alpha_t}} \Big( \mathbf{x}_t - \frac{1 - \alpha_t}{\sqrt{1 - \bar{\alpha}_t}} \mathbf{S}_\theta(\mathbf{x}_t, t) \Big)},
\end{split}
\end{equation}
and $\Sigma_{t}$ fixed as $\sqrt{\beta_{t}}$. 
Thus, the coefficients in Eq.~\ref{eq:RWCS_update} are $a=\frac{1}{\sqrt{\alpha_{t}}}$, $b=-\frac{1-\alpha_{t}}{\sqrt{\alpha_{t}(1 - \bar{\alpha}_t)}}$, and $c=\sqrt{\beta_{t}}$.

\subsection{Architectural Details}\label{sec:architecture}
The overall structure of \textsc{PDM} is illustrated in Fig.~\ref{fig:FIG_NETWORK}.
Here, we provide a more detailed explanation of each component in \textsc{PDM}.

\paragraph{Input Image Encoder} \(\mathcal{E}_\phi\) extracts multi-level features from \(I_s\) and \(I_d\) using a CNN-based structure, as described in \cite{sun2021loftr}:
\begin{equation}\label{eq:e_theta}
    (C_s, F_s) =  \, \mathcal{E}_\phi(I_s) 
    \, \text{and} \,
    (C_d, F_d) =  \, \mathcal{E}_\phi(I_d),
\end{equation}
where \(C_s\) and \(C_d\) are coarse-level features scaled by \(\times 1/8\), and \(F_s\) and \(F_d\) are fine-level features scaled by \(\times 1/2\).

\paragraph{Query Generator} includes a local feature amplifier \(\mathcal{E}_\rho\), based on \cite{4723207} and a keypoint detector $\mathcal{D}$ for generating the particles required for matching. 
We utilize SIFT~\cite{lowe2004distinctive} and Blob~\cite{ZhangWB15} detection algorithms on \(I_s\) to produce the query points \(\mathbf{x}^s\) for $\mathbf{S}_{\boldsymbol\theta}$, as described by:
\begin{equation}\label{eq:keypoint}
    \mathbf{x}^s = \mathcal{D}(\mathcal{E}_\rho(I_s)).
\end{equation}

\paragraph{Particle Initializer} generates random Gaussian points \(\mathbf{x}_T^d\) from a Gaussian distribution. 
\(\mathbf{x}_T^d\) has the same shape as the query points \(\mathbf{x}^s\) and is used by \(\mathbf{S}_\theta\) for subsequent refinement of the corresponding points.

\paragraph{RWCS Network} $\mathbf{S}_\theta$ takes a coarse-to-fine approach with two Random Walker Transformer (\texttt{RWTr}) networks following the ViT-based architecture~\cite{jiang2021cotr}, as shown in Fig.~\ref{fig:FIG_RWTR}.
$\mathbf{S}_\theta$ outputs the displacement \(\Delta \mathbf{x}_t^d = \mathbf{x}_t^d - \mathbf{x}_{t+1}^d\) using multi-level features: coarse-level features $C_s, C_d$ and fine-level features $F_s, F_d$ with higher spatial resolution:
\begin{equation}\label{eq:displacement_path}
\begin{split}
    \text{Coarse Level:} & \\
    \mathbf{\tilde{x}}_t^d = & \texttt{RWTr}(\mathbf{x}_{t+1}^d | C_s, C_d, \mathbf{x}^s, t), \\
    \text{Fine Level:} & \\
    \mathbf{x}_t^d = & \texttt{RWTr}(\mathbf{x}_{t+1}^d | F_s^{crop}, F_d^{crop}, \mathbf{x}^s, t), \\
\end{split}
\end{equation}
where $F_s^{crop} = \texttt{CROP}(F_s, \mathbf{x}^s)$ and $F_d^{crop} = \texttt{CROP}(F_d, \mathbf{\tilde{x}}_t^d)$ denote local patches cropped from the fine-level features centered at positions $\mathbf{x}^s$ and $\mathbf{\tilde{x}}_t^d$, respectively.

\subsection{Training Particle Diffusion Matching}\label{sec:train}
The loss function \(\mathcal{L}\) for end-to-end training of the \textsc{PDM} is defined as:
\begin{equation}\label{eq_full}
\begin{split}
\mathcal{L} = \mathcal{L}_{\text{D}} + \lambda_{\text{P}}\mathcal{L}_{\text{P}} + \lambda_{\text{R}}\mathcal{L}_{\text{R}},
\end{split}
\end{equation}
where \(\mathcal{L}_{\text{D}}\), \(\mathcal{L}_\text{P}\), and \(\mathcal{L}_\text{R}\) denote the DDPM loss, pixel matching loss, and regularization loss, respectively. 
\(\lambda_\text{P}\) and \(\lambda_{\text{R}}\) control the relative importance of the terms.
Each term, along with further details on the training process, will be described in the following subsections.

\subsubsection{DDPM Loss}
Following \citep{ho2020denoising}, we apply the simplified DDPM loss function \(L_{\text{D}}\) as:
\begin{equation}\label{eq:dsm_h}
\begin{split}
\mathcal{L}_{\text{D}} &= D_\text{KL}(q(\mathbf{x}^{d}_t \vert \mathbf{x}^{d}_{t+1}, \mathbf{x}_0^d) \parallel p_\theta(\mathbf{x}^{d}_t \vert\mathbf{x}^{d}_{t+1})) \\
 & = \mathbb{E}_{\mathbf{x}^{d}_{t}, t} \left\| \mathbf{S}_\theta(\mathbf{x}^{d}_t, t) - \epsilon \right\|_{2}^{2},
\end{split}
\end{equation}
where \(\epsilon \sim \mathcal{N}(0, \mathbf{I})\).

\subsubsection{Pixel Matching Loss} 
The pixel matching loss \(\mathcal{L}_{\text{P}}\) aims to maximize the similarity of aligned pixel appearances and is defined as~\cite{kim2022diffusemorph}:
\begin{equation}
\mathcal{L}_{\text{P}} = -\text{NCC}(\mathcal{B}(\hat{I}_{s}), \mathcal{B}(I_d)),
\label{L_L}
\end{equation}
where \text{NCC} denotes the normalized cross-correlation between the images, \(\hat{I}_s\) is the warped image from \(I_s\) using a second-order polynomial transformation~\cite{10_1007} based on the matches between \(\mathbf{x}^s\) and \(\mathbf{x}_0^d\), and \(\mathcal{B}\) is an image enhancement filter~\cite{BahadarKhan2016morph} used to generate simplified binary images.

\subsubsection{Regularization Loss}
To complement the DDPM loss in Eq.~\ref{eq:dsm_h}, we define a regularization term to ensure that all particles converge to their appropriate positions:
\begin{equation}
    \mathcal{L}_{\text{R}} = \left\lVert  H - H_{gt} \right\rVert_1,
    \label{eq:loss_reg}
\end{equation}
where \(H_{gt}\) is the ground truth homography, and \(H\) is the estimated homography obtained by a linear optimization algorithm using differentiable RANSAC~\cite{Brachmann2019DSACstar} with a threshold of $2$, based on the correspondences between \(\mathbf{x}^s\) and \(\mathbf{x}_0^d\).

\section{Experiments}
\subsection{Datasets}\label{subsec:dataset}
We evaluated \textsc{PDM} across three dataset types: SFI-UWFI pairs captured under varying conditions, SFI-SFI pairs with different angles, and UWFI-UWFI pairs from fluorescein angiography for montage tasks.

For SFI-UWFI evaluation, we used a dataset from Kangbuk Samsung Medical Center (KBSMC)\footnote{This study adhered to the tenets of the Declaration of Helsinki and was approved by the Institutional Review Boards (IRB) of Kangbuk Samsung Hospital (No. KBSMC 2019-08-031). The study is a retrospective review of medical records, and the data were fully anonymized prior to processing. The IRB waived the requirement for informed consent.} comprising 3,744 SFI-UWFI pairs, where SFIs are approximately \(1\times \sim 4\times\) smaller than UWFIs but captured from the same patients. 
The dataset was split into 3,370 training and 374 test pairs. We manually generated ground truth points correspondence and trained the model in a fully supervised manner.
For SFI-SFI and UWFI-UWFI, we used the FIRE~\cite{hernandezmatas2017fire} and FLORI21~\cite{Li2021Flori} datasets, respectively.

\subsection{Implementation Details}\label{subsec:implementation}
\paragraph{Common setup}
All images from KBSMC, FIRE~\cite{hernandezmatas2017fire} and FLORI21~\cite{Li2021Flori} were resized to \(768 \times 768\) resolution for training and testing. 
Training was run for \(\ge 1K\) epochs with an NVIDIA RTX 4090, and AdamW optimizer with initial learning rate \(0.001\), \(\beta_1 = 0.9\), \(\beta_2 = 0.999\), and \(\epsilon = 10^{-8}\), applying weight decay every \(100\)K iterations with a decay rate \(0.01\)
and learning rate halving every \(150\)K iterations. 
We set \(\lambda_{\text{P}}\) and \(\lambda_{\text{R}}\) to \(0.1\) and \(0.5\), respectively. 
Number of particles were set to 100.
Data augmentation included random rotations (90$^{\circ}$, 180$^{\circ}$, or 270$^{\circ}$) and photometric distortions (illumination, contrast, Gaussian blur, and Gaussian noise). 
Point coordinates were normalized in the range \([-1, 1]\).
We set sampling steps \(T\) for training and testing the \textsc{PDM} to \(100\).

\paragraph{Input Image Encoder}
To extract multi-scale features, we utilized the ResNet~\cite{he2015deepresiduallearningimage}-based encoder structure from LoFTR~\cite{sun2021loftr}.

\paragraph{Query Generator}
To facilitate the detection of feature points in the image, we first applied a filtering-based operation~\cite{4723207} as a local feature amplifier \(\mathcal{E}_\rho\) to the source image. 
Subsequently, to generate 100 query points, we initially used OpenCV's SIFT~\cite{lowe2004distinctive} key point extraction algorithm. 
In cases where the number of points was insufficient, we applied a Blob~\cite{ZhangWB15} detection algorithm.

\paragraph{Particle Initializer}
Once the query points set is finalized, a noisy points set with the same shape as the query points set is extracted from the particle initializer using a Gaussian distribution. 
This is achieved using the \texttt{randn()} function from the PyTorch library.

\paragraph{RWCS Network}
The components of \texttt{RWTr} include a multi-scale encoder, transformer layers, and an MLP.

In the multi-scale encoder, the feature map from the Input Image Encoder $\mathcal{E}_\phi$ is flattened and passed through a linear layer to be encoded into a final $512$-dimensional feature vector.

The transformer layers consist of $16$ transformer blocks, each featuring a transformer encoder layer with an input vector size of $512$, $4$ attention heads, a feed-forward dimension of $2048$, and a dropout rate of $0.1$.

Finally, the MLP is composed of three linear layers: an input layer with a dimension of $512$, a hidden layer with $128$ dimensions, and an output layer with a dimension of $2$. 
The activation function used is \texttt{ReLU}.

\subsection{Comparative Evaluation Settings}\label{subsec:baseline}
Baselines include SuperPoint~\cite{detone2018superpoint}, GLAMpoints~\cite{truong2019glampoints}, ISTN~\cite{lee2019istn}, NCNet~\cite{rocco2020ncnet}, SuperGlue~\cite{sarlin2020superglue}, REMPE~\cite{hernandez2020rempe}, DLKFM~\cite{zhao2021deep}, LoFTR~\cite{sun2021loftr}, IHN~\cite{cao2022iterative}, SuperRetina~\cite{liu2022semi}, ASPanFormer~\cite{chen2022aspanformer}, GeoFormer~\cite{liu2023geometrized}, MCNet~\cite{zhu2024mcnet}, XoFTR~\cite{tuzcuouglu2024xoftr}, RetinaRegNet~\cite{sivaraman2024retinaregnetzeroshotapproachretinal}, ROMA~\cite{edstedt2024roma}, and MINIMA~\cite{ren2025minima} with ROMA backbone~\cite{edstedt2024roma}.

All comparative methods were trained separately for each dataset evaluation, except for RetinaRegNet~\cite{sivaraman2024retinaregnetzeroshotapproachretinal}, which is an inference-only method that requires no training.
For evaluation on FIRE~\cite{hernandezmatas2017fire}, methods including PDM were trained in a self-supervised manner using SFIs from KBSMC and FIRE, with warped pairs synthesized via random homographies.
For FLORI21~\cite{Li2021Flori}, methods including \textsc{PDM} were similarly trained using random homography-warped UWFI pairs from KBSMC.

To ensure statistical rigor, we report the mean and standard deviation of \textsc{PDM} from $10$ independent runs with different random initializations.

We use the landmark point-based CEM approach~\cite{truong2019glampoints,liu2022semi, liu2023geometrized}, with quantitative measurement of the median error (MEE) and maximum error (MAE), as well as the Area Under Curve (AUC)~\cite{hernandezmatas2017fire}, and the mean on all image pairs, mAUC.
We also categorized the results as: 
i) \textit{Failed} (no transform created),
ii) \textit{Acceptable} (MAE \(<\) $50$ and MEE \(<\) $20$),
iii) \textit{Inaccurate} (all others).

\begin{figure*}[!htbp] 
    \centering
    \includegraphics[width=1.0\textwidth]{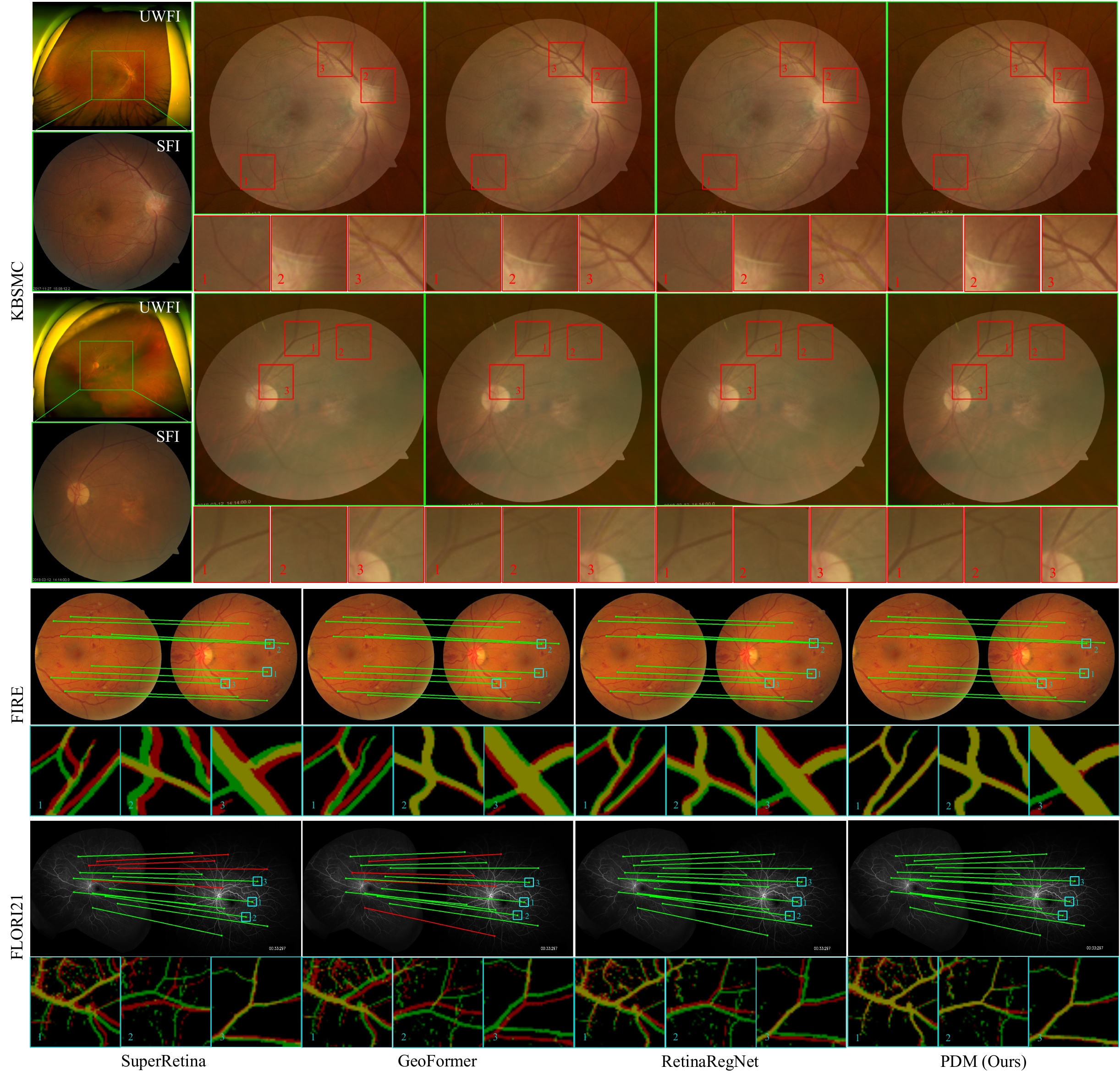}
    \caption{
    \textbf{Qualitative evaluation of \textsc{PDM} on the KBSMC, FIRE, and FLORI21 datasets.}
    For input image pairs with significantly different styles, \textsc{PDM} robustly identify sufficient matches compared to SuperRetina~\cite{liu2022semi},  GeoFormer~\cite{liu2023geometrized} and RetinaRegNet~\cite{sivaraman2024retinaregnetzeroshotapproachretinal}.
    The red box indicates the zoomed-in overlay region between SFI and UWFI for the KBSMC dataset. 
    For the FIRE and FLORI21 datasets, correspondence matches are visualized with green lines (accurate matches) and red lines (inaccurate matches), while the blue boxes show magnified regions for verifying vessel alignment.
    } 
    \label{fig:FIG_ALL}
\end{figure*}

\begin{table*}[!htbp]
\renewcommand*{\arraystretch}{1.1}
\caption {Quantitative evaluation on KBSMC, FIRE, and FLORI21 dataset.}
\label{tab:ALL}
\resizebox{1.0\textwidth}{!}{
\begin{tabular}{lcccccccc}
\hline
\multirow{2}{*}{\textbf{Methods}}                                                          
&   \multicolumn{2}{c}{\textbf{KBSMC}}                                          
& & \multicolumn{2}{c}{\textbf{FIRE}}                                             
& & \multicolumn{2}{c}{\textbf{FLORI21}}   \\ \cline{2-3} \cline{5-6} \cline{8-9} 
& \textit{Acceptable}(\%, $\uparrow$)      & \textbf{mAUC} ($\uparrow$)                           & 
& \textit{Acceptable}(\%, $\uparrow$)      & \textbf{mAUC} ($\uparrow$)                           &  
& \textit{Acceptable}(\%, $\uparrow$)      & \textbf{mAUC} ($\uparrow$)                              \\ \hline
SuperPoint~\cite{detone2018superpoint}                               & 9.09$^{*}$                               & 8.7$^{*}$                                     &           & 94.78$^{*}$                                    & 67.3$^{*}$                               &  & 40$^{*}$                                 & 39.1$^{*}$                              \\
GLAMpoints~\cite{truong2019glampoints}                               & 9.89$^{*}$                               & 8.4$^{*}$                                     &           & 93.28$^{*}$                                    & 61.9$^{*}$                               &  & 33.33$^{*}$                              & 34.4$^{*}$                              \\
ISTN~\cite{lee2019istn}                                              & 20.86$^{*}$                              & 12.1$^{*}$                                    &           & 86.57$^{*}$                                    & 60.9$^{*}$                               &  & 53.33$^{*}$                              & 52.5$^{*}$                              \\
NCNet~\cite{rocco2020ncnet}                                          & 12.30$^{*}$                              & 9.6$^{*}$                                     &           & 86.57$^{*}$                                    & 61.4$^{*}$                               &  & 53.33$^{*}$                              & 50.8$^{*}$                              \\
SuperGlue~\cite{sarlin2020superglue}                                 & 24.06$^{*}$                              & 15.3$^{*}$                                    &           & 95.52$^{*}$                                    & 68.7$^{*}$                               &  & 80$^{***}$                                 & 59.8$^{*}$                              \\
REMPE~\cite{hernandez2020rempe}                                      & 22.46$^{*}$                              & 15.0$^{*}$                                    &           & 97.01$^{**}$                                    & 72.1$^{*}$                               &  & 73.33$^{**}$                              & 60.0$^{*}$                              \\
DLKFM~\cite{zhao2021deep}                                            & 22.73$^{*}$                              & 13.5$^{*}$                                    &           & 86.57$^{*}$                                    & 61.4$^{*}$                               &  & 40$^{*}$                                 & 40.1$^{*}$                              \\
LoFTR~\cite{sun2021loftr}                                            & 26.20$^{*}$                              & 16.9$^{*}$                                    &           & 97.01$^{**}$                                    & 71.5$^{*}$                               &  & 66.67$^{*}$                              & 51.5$^{*}$                              \\
COTR~\cite{jiang2021cotr}                                            & 32.62$^{*}$                              & 17.2$^{*}$                                    &           & 97.01$^{**}$                                    & 70.8$^{*}$                               &  & 60$^{*}$                              & 49.8$^{*}$                              \\
IHN~\cite{cao2022iterative}                                          & 23.80$^{*}$                              & 14.5$^{*}$                                    &           & 88.81$^{*}$                                    & 63.5$^{*}$                               &  & 60$^{*}$                                 & 50.0$^{*}$                              \\
SuperRetina~\cite{liu2022semi}                                       & 34.76$^{*}$                              & 22.3$^{*}$                                    &           & 98.51$^{\text{ns}}$       & 75.5$^{**}$                               &  & 80$^{***}$                                 & 65.0$^{*}$                              \\
ASPanFormer~\cite{chen2022aspanformer}                               & 24.87$^{*}$                              & 16.2$^{*}$                                    &           & 92.54$^{*}$                                    & 70.4$^{*}$                               &  & 73.33$^{**}$                              & 62.8$^{*}$                              \\
GeoFormer~\cite{liu2023geometrized}                                  & 36.10$^{*}$ & 24.1$^{*}$       &           & 98.51$^{\text{ns}}$       & 75.6$^{**}$                               &  & 93.33$^{\text{ns}}$ & 71.4$^{*}$                              \\ 
MCNet~\cite{zhu2024mcnet}                                            & 32.89$^{*}$                              & 20.9$^{*}$                                    &           & 92.54$^{*}$                                    & 69.3$^{*}$                               &  & 60$^{*}$                                 & 48.6$^{*}$                              \\
XoFTR~\cite{tuzcuouglu2024xoftr}                                            & 35.29$^{*}$                              & 23.9$^{*}$                                    &           & 98.51$^{\text{ns}}$                                    & 75.7$^{**}$                               &  & 93.33$^{\text{ns}}$                                 & 75.3$^{*}$                              \\
RetinaRegNet~\cite{sivaraman2024retinaregnetzeroshotapproachretinal} & 31.28$^{*}$                              & 20.3$^{*}$                                    &           & \textbf{99.25}$^{\text{ns}}$          & 77.9$^{\text{ns}}$ &  & \textbf{100}$^{\text{ns}}$      & 86.8$^{***}$ \\ 
ROMA~\cite{edstedt2024roma}                                            & 33.42$^{*}$                              & 23.1$^{*}$                                    &           & 98.51$^{\text{ns}}$                                    & 75.3$^{**}$                               &  & 93.33$^{\text{ns}}$                                 & 73.4$^{*}$                              \\
MINIMA~\cite{ren2025minima,edstedt2024roma}                                            & \underline{40.91}$^{*}$                              & \underline{27.6}$^{*}$                                    &           & \textbf{99.25}$^{\text{ns}}$                                    & \underline{76.9}$^{***}$                               &  & \textbf{100}$^{\text{ns}}$                                 & \underline{91.4}$^{\text{ns}}$                              \\ \hline
\textsc{PDM} (ours)                                                        & \textbf{58.56±2.1}    & \textbf{34.8±1.3} & \textbf{} & \textbf{99.25±0.4} & \textbf{78.2±0.9}     &  & \textbf{100±0.0}      & \textbf{92.3±1.2}    \\ 
 \hline
\end{tabular}}
\scriptsize{
The bold and underline values denote the best and second best results, respectively. \\
Standard deviations are reported for PDM from 10 independent runs.\\
Statistical significance (Wilcoxon rank-sum test vs. PDM): $^{*}$$p<0.001$, $^{**}$$p<0.01$, $^{***}$$p<0.05$, and $^{\text{ns}}$not significant.\\
The \textit{Failed} rate for all methods across all datasets is $0\%$, and the \textit{Inaccurate} rate is the percentage excluding the \textit{Acceptable} rate.}
\end{table*}

\begin{table}[!htbp]
\renewcommand*{\arraystretch}{1.2}
\caption{Computational cost comparison}
\resizebox{1.0\columnwidth}{!}{
\begin{tabular}{lccc}
\hline
\textbf{Methods} & \textbf{Inference time} & \textbf{Memory (model / inference)} \\
\hline
MCNet~\cite{zhu2024mcnet}      & 35.76s  & 1692MB / 2280MB \\
GeoFormer~\cite{liu2023geometrized}  & 1.15s  & 242MB / 918MB \\
RetinaRegNet~\cite{sivaraman2024retinaregnetzeroshotapproachretinal}     & 0.09s  & 4826MB / 5415MB \\
\textsc{PDM} (ours)     & 0.45s  & 1132MB / 1721MB \\
\hline
\end{tabular}}
\label{tab:runtime_memory}
\end{table}

\subsection{Performance Evaluation}\label{subsec:eval}

Tab.~\ref{tab:ALL} presents quantitative comparative evaluation results, where it achieves the highest performance in both the \textit{Acceptable} rate and mAUC metrics on KBSMC, FIRE\cite{hernandezmatas2017fire}, and FLORI21~\cite{Li2021Flori} datasets. 
We also conducted Wilcoxon rank-sum tests~\cite{inbookHaynes} to assess the statistical significance of performance differences using \(p\)-value between PDM and baseline methods,
Notably, \textsc{PDM} outperforms all competitors, achieving a $22.46\%$ higher \textit{Acceptable} rate and a $10.7\%$ increase in mAUC over GeoFormer, demonstrating its superior alignment capabilities on KBSMC dataset.
Even compared to the recent MINIMA~\cite{ren2025minima,roma}, which claims modality-invariant matching capabilities, PDM shows substantial improvements with $17.65\%$ higher \textit{Acceptable} rate and $7.2\%$ increase in mAUC, demonstrating superior cross-modal alignment capabilities on the challenging SFI-UWFI task.
Fig.~\ref{fig:FIG_ALL} further highlights the effectiveness of \textsc{PDM} compared to state-of-the-art methods.
Also, we visualize keypoint correspondences from \textsc{PDM} and state-of-the-art methods, including SuperRetina~\cite{liu2022semi}, GeoFormer~\cite{liu2023geometrized}, and RetinaRegNet~\cite{sivaraman2024retinaregnetzeroshotapproachretinal}. 
Using GT source key points, destination points from each method are evaluated, with green lines indicating correct matches and red lines representing errors. 
The zoomed-in boxes highlight local improvements of \textsc{PDM}, with vessel alignment shown for visualization.
To enhance alignment on FIRE, \textsc{PDM} was trained in a self-supervised manner using synthetic homographies and image pairs from the SFI in KBSMC and FIRE datasets. 
Similarly, for FLORI21, synthetic transformations from the UWFI in KBSMC and FLORI21 datasets were used, enabling \textsc{PDM} to achieve superior alignment accuracy and robustness.

Tab.~\ref{tab:runtime_memory} presents computational comparison. 
Although \textsc{PDM} adopts an iterative correspondence refinement strategy, it achieves a notably low inference time of 0.45 seconds. 
This efficiency stems from performing the refinement process only on a sparse set of point correspondences, rather than over dense grids. 
Furthermore, the memory consumption during inference (1721MB) remains reasonable and well-balanced, demonstrating that our method provides a favorable trade-off between performance and computational cost compared to existing approaches.

\begin{table}[!htbp]
\renewcommand*{\arraystretch}{1.1}
\caption {Ablation study of different particle configurations.}
\label{tab:ablative}
\resizebox{1.0\columnwidth}{!}{
{\tiny
\begin{tabular}{ccccc}
\hline
\multirow{2}{*}{\textbf{Particle Count}} & 
\multicolumn{3}{c}{\textbf{mAUC} ($\uparrow$)} \\ \cline{2-4} 
& \textbf{KBSMC} & \textbf{FIRE} & \textbf{FLORI21} \\ \hline
4      & 24.2±1.8  & 65.8±1.5  & 71.3±1.9    \\
50     & 34.5±1.2  & 76.2±1.1  & 85.2±1.4    \\
100    & \textbf{34.8±1.3}  & \textbf{78.2±0.9}  & 92.3±1.2    \\
200    & \textbf{34.8±1.2}  & 78.1±0.8  & \textbf{92.4±1.1}    \\
500    & \textbf{34.8±1.1}  & \textbf{78.2±0.9}  & 92.2±1.0    \\ \hline
\end{tabular}}}
\scriptsize{The bold values denote the best results. \\
All values are reported as mean±std from 10 independent runs.}
\end{table}

\begin{table}[!htbp]
\renewcommand*{\arraystretch}{1.2}
\caption {Ablation study of RWCS Network Variants.}
\label{tab:ctf}
\resizebox{1.0\columnwidth}{!}{
{\tiny
\begin{tabular}{clccc}
\hline
\multicolumn{2}{c}{\multirow{2}{*}{\textbf{RWCS Network Configurations}}} & \multicolumn{3}{c}{\textbf{mAUC} ($\uparrow$)} \\ \cline{3-5} 
\multicolumn{2}{c}{}                                              & \textbf{KBSMC} & \textbf{FIRE} & \textbf{FLORI21} \\ \hline
\multicolumn{2}{c}{w/ Coarse-Level \texttt{RWTr}}                                  & 32.1±1.4   & 74.2±1.2 & 88.6±1.3     \\
\multicolumn{2}{c}{w/ Fine-Level \texttt{RWTr}}                                  & 32.9±1.5   & 75.0±1.1 & 90.8±1.2     \\
\multicolumn{2}{c}{w/ Coarse-to-Fine \texttt{RWTr}s}                                & \textbf{34.8±1.3}   & \textbf{76.2±0.9} & \textbf{92.3±1.2}     \\ \hline
\end{tabular}}}
\scriptsize{The bold values denote the best results. \\
All values are reported as mean±std from 10 independent runs.}
\end{table}

\begin{table}[!htbp]
\renewcommand*{\arraystretch}{1.1}
\caption {Ablation study of different query generator.}
\label{tab:query}
\resizebox{1.0\columnwidth}{!}{
{\tiny
\begin{tabular}{lcccc}
\hline
\multirow{2}{*}{\textbf{Query Generators}} & 
\multicolumn{3}{c}{\textbf{mAUC} ($\uparrow$)} \\ \cline{2-4} 
& \textbf{KBSMC} & \textbf{FIRE} & \textbf{FLORI21} \\ \hline
SIFT~\cite{lowe2004distinctive} + Blob~\cite{ZhangWB15}      & \textbf{34.8±1.3}  & 78.2±0.9  & 92.3±1.2    \\
SURF~\cite{bay2006surf}     & 34.3±1.4  & 77.9±1.0  & 91.9±1.3    \\
FAST~\cite{RostenDrummond2006}    & 32.0±1.5  & 75.1±1.2  & 87.3±1.4    \\
Haris Corner Detector~\cite{Harris88alvey}    & 30.9±1.6  & 72.3±1.3  & 85.8±1.5    \\
KAZE~\cite{Alcantarilla2012KAZE}    & 34.7±1.3  & 78.1±0.9  & \textbf{92.4±1.1}    \\ 
SuperPoint~\cite{detone2018superpoint}  & 33.5±1.1  & \textbf{78.3±1.1}  & 90.6±1.2                                 \\ 
Random Initialization  & 29.6±1.8  & 70.3±1.6  & 83.0±1.7   \\ \hline
\end{tabular}}}
\scriptsize{The bold values denote the best results. \\
All values are reported as mean±std from 10 independent runs.}
\end{table}

\begin{table}[!htbp]
\renewcommand*{\arraystretch}{1.1}
\caption {Ablation study of different particle initialization.}
\label{tab:particle}
\resizebox{1.0\columnwidth}{!}{
{\tiny
\begin{tabular}{lcccc}
\hline
\multirow{2}{*}{\textbf{Particle Range}} & 
\multicolumn{3}{c}{\textbf{mAUC} ($\uparrow$)} \\ \cline{2-4} 
& \textbf{KBSMC} & \textbf{FIRE} & \textbf{FLORI21} \\ \hline
\([-0.1, 0.1]\)      & 30.0±1.5  & 73.1±1.3  & 84.9±1.4    \\
\([-1, 1]\)     & \textbf{34.8±1.3}  & \textbf{78.2±0.9}  & \textbf{92.3±1.2}    \\
\([-10, 10]\)    & 33.2±1.4  & 76.0±1.1  & 88.8±1.3    \\
\([-100, 100]\)    & 25.4±1.7  & 65.1±1.5  & 70.9±1.6    \\
\([-384, 384]\) (full resolution)    & 23.1±1.9  & 61.8±1.7  & 64.6±1.8    \\  \hline
\end{tabular}}}
\scriptsize{The bold values denote the best results. \\
All values are reported as mean±std from 10 independent runs.}
\end{table}

\begin{table}[!htbp]
\renewcommand*{\arraystretch}{1.1}
\caption {Ablation study of different sampling steps.}
\label{tab:sampling}
\resizebox{1.0\columnwidth}{!}{
{\tiny
\begin{tabular}{ccccc}
\hline
\multirow{2}{*}{\textbf{Sampling Steps}} & 
\multicolumn{3}{c}{\textbf{mAUC} ($\uparrow$)} \\ \cline{2-4} 
& \textbf{KBSMC} & \textbf{FIRE} & \textbf{FLORI21} \\ \hline
50      & 34.2±1.4  & 77.8±1.0  & 91.5±1.3    \\
100     & \textbf{34.8±1.3}  & \textbf{78.2±0.9}  & \textbf{92.3±1.2}    \\
250    & 34.7±1.3  & \textbf{78.2±0.9}  & 92.2±1.2    \\
500    & 33.2±1.5  & 76.9±1.1  & 90.8±1.4    \\
1000    & 30.8±1.6  & 75.9±1.3  & 89.4±1.5    \\ \hline
\end{tabular}}}
\scriptsize{The bold values denote the best results. \\
All values are reported as mean±std from 10 independent runs.}
\end{table}

\begin{table}[!htbp]
\renewcommand*{\arraystretch}{1.1}
\caption {Ablation study of zero-shot evaluation.}
\label{tab:zero}
\resizebox{1.0\columnwidth}{!}{
{\tiny
\begin{tabular}{lcccc}
\hline
\multirow{2}{*}{\textbf{Methods}} & 
\multicolumn{3}{c}{\textbf{mAUC} ($\uparrow$)} \\ \cline{2-4} 
& \textbf{KBSMC} & \textbf{FIRE} & \textbf{FLORI21} \\ \hline
GeoFormer~\cite{liu2023geometrized}     & 24.1  & 70.3  & 65.9    \\
ROMA~\cite{edstedt2024roma}    & 23.1  & 73.3  & 69.42    \\
MINIMA~\cite{ren2025minima,edstedt2024roma}    & 27.6  & 75.1  & 87.5    \\
PDM (ours)    & \textbf{34.8±1.3}  & \textbf{75.8±1.2}  & \textbf{87.9±1.3}    \\ \hline
\end{tabular}}}
\scriptsize{The bold values denote the best results. \\
All values are reported as mean±std from 10 independent runs.}
\end{table}

\subsection{Ablation Study}\label{subsec:ablation}

\paragraph{Particle Count}
We conducted an ablation study to evaluate the effect of particle count on the performance of \textsc{PDM}. 
The number of particles was varied from $4$ to $500$ to explore the trade-off between alignment accuracy and computational efficiency. 
As shown in Tab.~\ref{tab:ablative}, performance (mAUC) improved as the number of particles increased, but gains saturated beyond $100$ for most datasets. 
Using only $4$ particles resulted in poor alignment as they failed to adequately capture complex structural variations. 
In  contrast, increasing the particle count beyond $100$ offered minimal additional benefit, indicating diminishing returns. 
These findings indicate that using $100$ particles achieves an optimal balance between alignment accuracy and computational efficiency.
Notably, the computational overhead from varying particle count is negligible.

\paragraph{Random Walker Variants}
The RWCS Network \(\mathbf{S}_\theta\) includes two primary variants: the coarse-level \texttt{RWTr} and the coarse-to-fine cascade \texttt{RWTr}s. 
These variants reflect a trade-off between computation cost and alignment performance.

To further examine this trade-off, we evaluated the performance of each variant using the mAUC metric across three datasets: KBSMC, FIRE~\cite{hernandezmatas2017fire}, and FLORI21~\cite{Li2021Flori}. 
By comparing the results, we aim to assess how each variant balances computational efficiency and alignment accuracy across different scenarios.

Tab.~\ref{tab:ctf} summarizes the quantitative results for the various RWCS Network configurations.
The results indicate that the Coarse-to-Fine \texttt{RWTr}s variant consistently outperforms both single-scale variants across all datasets. While Fine-Level \texttt{RWTr} shows marginal improvements over Coarse-Level \texttt{RWTr}, neither single-scale approach matches the performance of our multi-scale strategy.
Notably, since all variants share the same network architecture, the memory footprint remains identical and runtime differences are negligible, as the Coarse-to-Fine approach simply applies the shared network twice sequentially. This design enables superior accuracy without the significant computational penalties typically associated with multi-scale methods, making it particularly suitable for applications prioritizing registration quality.

\paragraph{Query Generator Variants}
The query generator in \textsc{PDM} detects keypoints from the source SFI and uses them as conditions for the diffusion random walk.
Since SFIs are relatively cleaner and richer in texture compared to UWFIs, keypoint detection is more straightforward in this modality.
Tab.~\ref{tab:query} compares various query point initialization strategies, including both handcrafted keypoint detectors and random sampling methods. 
Interestingly, methods such as SIFT~\cite{lowe2004distinctive}, SURF~\cite{bay2006surf}, FAST~\cite{RostenDrummond2006}, KAZE~\cite{Alcantarilla2012KAZE}, and SuperPoint~\cite{detone2018superpoint} pretrained on each datasets (KBSMC, FIRE~\cite{hernandezmatas2017fire}, and FLORI21~\cite{Li2021Flori}) exhibit only marginal differences in performance, while even randomly initialized query points yield reasonably competitive results. 
These results suggest that our model is robust to the specific choice of keypoint detector. 
However, the spatial distribution of query points plays a crucial role in performance. 
When keypoints are well-distributed across the entire image and capture the underlying overall structure (e.g., SIFT~\cite{lowe2004distinctive} + Blob~\cite{ZhangWB15}), the model consistently achieves better results. 
In contrast, detectors that produce spatially biased or locally clustered keypoints (e.g., Harris Corner~\cite{Harris88alvey}) tend to result in suboptimal performance, highlighting the importance of globally informative initialization.

\paragraph{Particle Initialization}  
Tab.~\ref{tab:particle} presents an ablation study on different initialization ranges for the correspondence points ${\mathbf{x}^{s}}$ and ${\mathbf{x}^{d}}$. 
In our default setting, coordinates are normalized to the range \([-1, 1]\), which aligns well with the sampling behavior of the proposed diffusion model. 
To better illustrate the diffusion random walk process, Fig.~\ref{fig:FIG_TEASER} uses \([-10, 10]\) particle normalization instead of the standard \([-1, 1]\) range.
When the initialization range is too narrow (e.g., \([-0.1, 0.1]\)), the particle distribution becomes overly concentrated and resembles Gaussian noise. 
Although this may align with theoretical assumptions of the diffusion process, it can lead to instability during denormalization to full-resolution coordinates, as small perturbations can result in disproportionately large displacements. 
Conversely, using a range that is too broad (e.g., \([-100, 100]\) or full-resolution \([-384, 384]\)) causes particles to be initialized far from the target distribution. 
This makes it challenging for the model to effectively denoise within a fixed number of sampling steps (e.g., 100), resulting in degraded performance due to a mismatch between training and inference conditions. 
These findings underscore the importance of selecting an appropriate initialization range to balance the expressiveness of the particle space and the denoising capacity of the diffusion process.

\paragraph{Sampling Steps}  
Tab.~\ref{tab:sampling} examines the effect of the number of sampling steps in our diffusion-based alignment framework. 
When the number of steps is too low (e.g., 50), the model fails to adequately refine the randomly initialized particles, resulting in suboptimal alignment performance. 
Conversely, increasing the number of steps beyond a certain threshold (e.g., 500 or 1000)  leads to performance degradation. 
We hypothesize that this decline stems from the nature of diffusion models: excessive iterations may produce unintended or over-smoothed features that hinder accurate alignment~\cite{aithal2024understandinghallucinationsdiffusionmodels}. 
Based on these findings, we choose 100 sampling steps as a balanced setting, providing sufficient refinement while maintaining stable alignment performance across all benchmarks.

\paragraph{Generalization}  
As shown in Tab.~\ref{tab:zero}, all methods were trained on the KBSMC dataset and evaluated on FIRE~\cite{hernandezmatas2017fire} and FLORI21~\cite{Li2021Flori} in a zero-shot setting, without any fine-tuning or domain-specific adaptation.
The proposed PDM achieves consistently superior performance across unseen datasets, demonstrating its strong generalization capability and robustness beyond the training domain.

\begin{figure}[!htbp] 
    \centering
    \includegraphics[width=1.0\columnwidth]{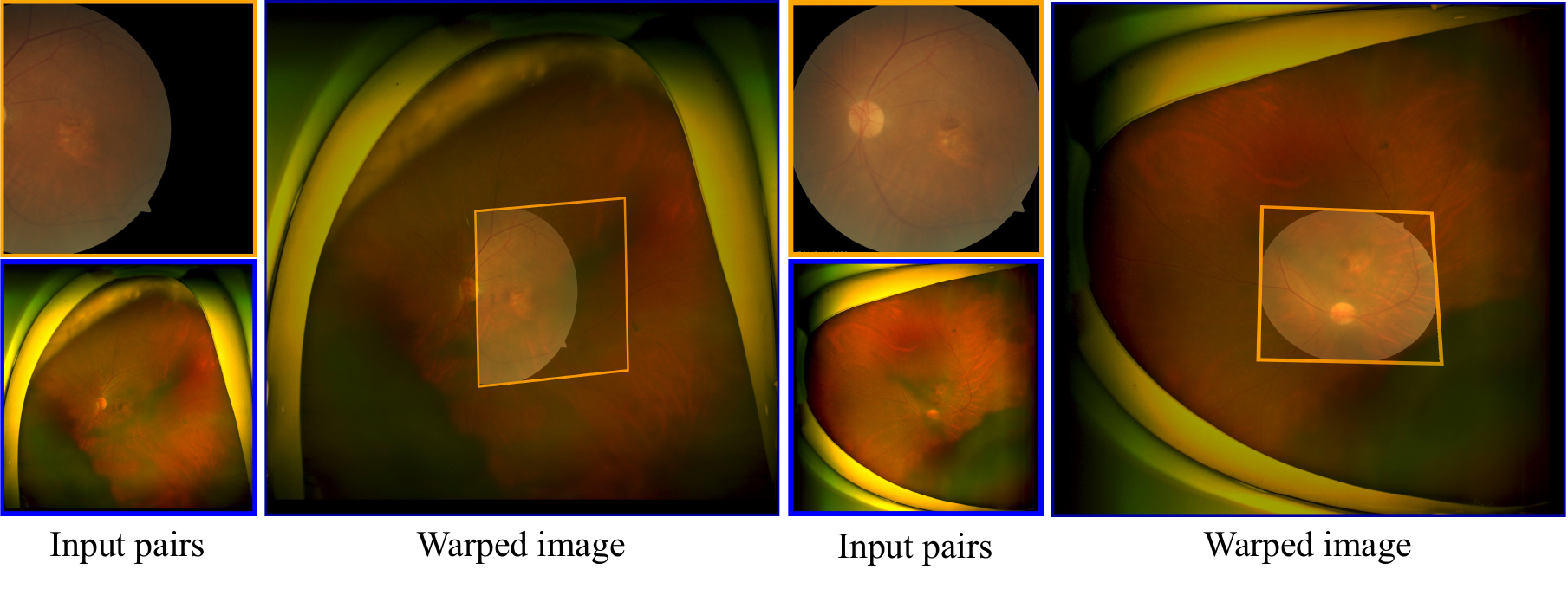}
    \caption{\textbf{Extreme cases.} The transformation estimation results of \textsc{PDM} are presented on synthesized less-overlapping pair (left) and rotated pair (right).
    }
    \label{fig:FIG_EXTR1}
\end{figure}

\begin{figure}[!htbp] 
    \centering
    \includegraphics[width=1.0\columnwidth]{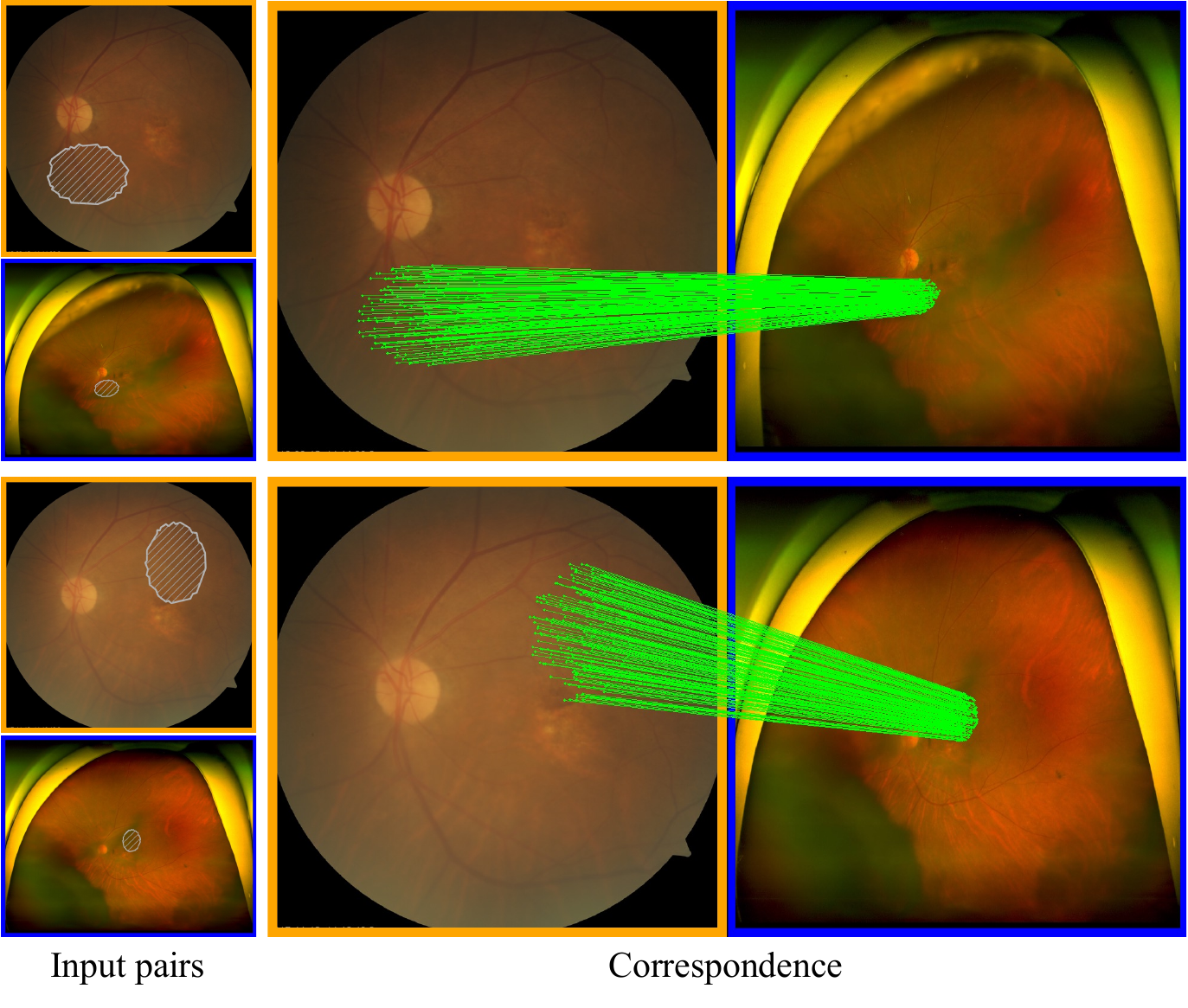}
    \caption{\textbf{Partial correspondence.} The transformation estimation results of \textsc{PDM} are presented on two intentionally selected textureless regions to demonstrate robustness.
    }
    \label{fig:FIG_EXTR2}
\end{figure}

\begin{figure}[!htbp] 
    \centering
    \includegraphics[width=1.0\columnwidth]{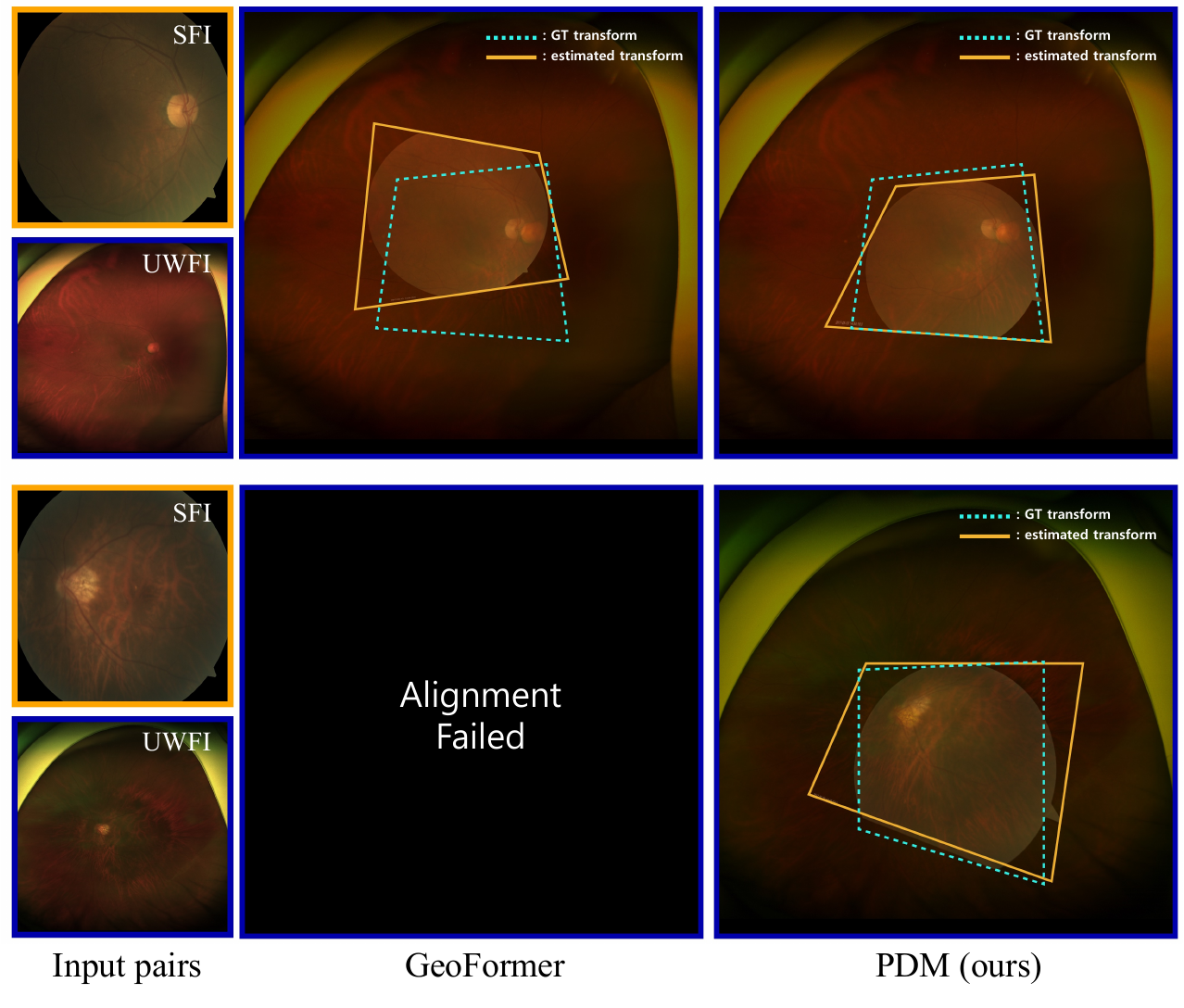}
    \caption{\textbf{Failure cases.} The transformation estimation results of \textsc{PDM} and the baseline GeoFormer~\cite{liu2023geometrized} are presented on two highly challenging registration samples from the KBSMC dataset.
    }
    \label{fig:FIG_FAIL}
\end{figure}

\begin{figure}[!htbp] 
    \centering
    \includegraphics[width=1.0\columnwidth]{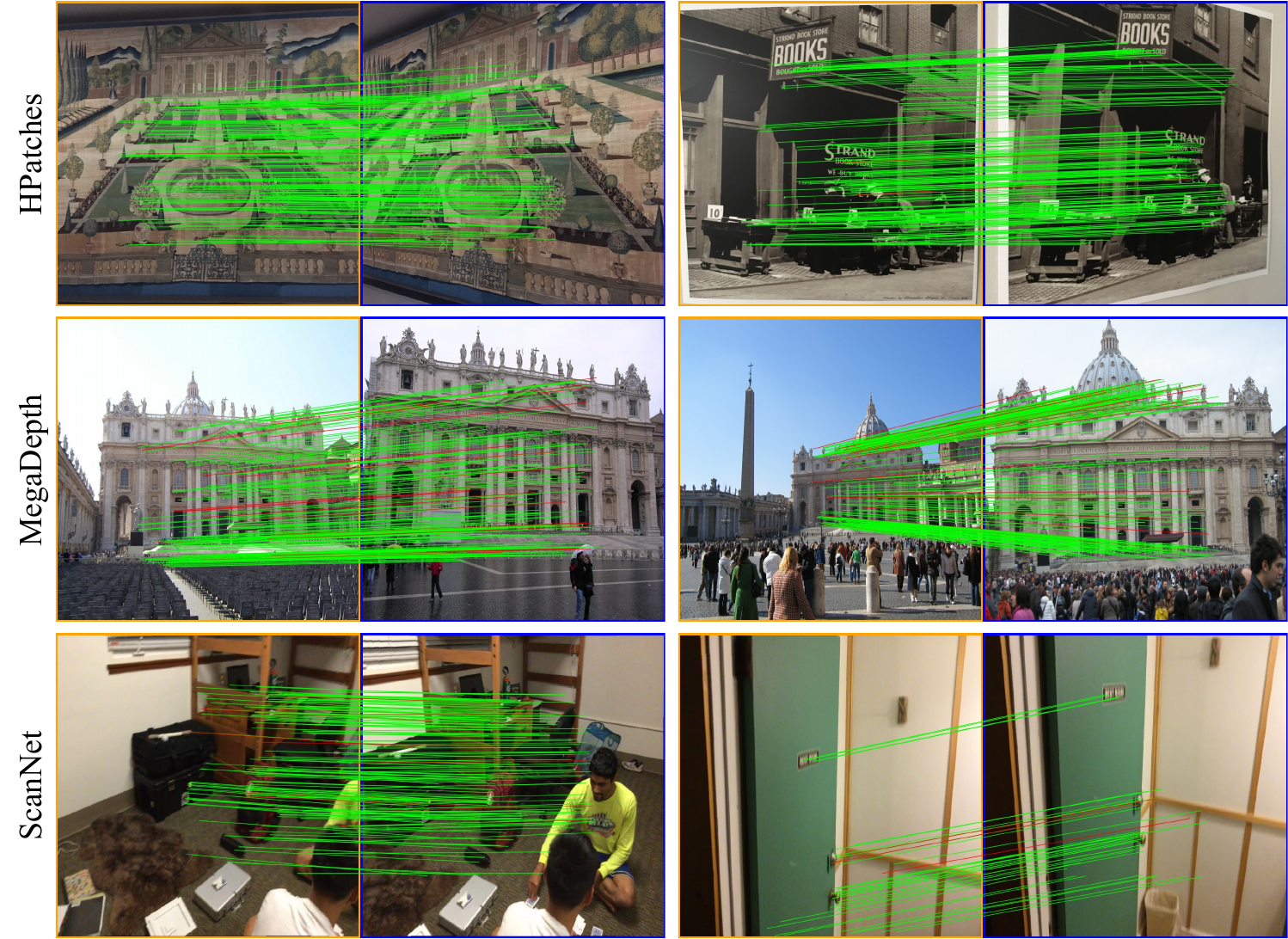}
    \caption{\textbf{Qualitative evaluation of \textsc{PDM} on the HPatches, MegaDepth, and ScanNet datasets.} Correspondence matches are visualized with green lines (accurate matches) and red lines (inaccurate matches).
    }
    \label{fig:FIG_HP}
\end{figure}

\begin{table}[!htbp]
{
\caption {\textbf{Quantitative Evaluation on Scene Datasets.}}
\label{tab:scenery}
\resizebox{1.0\columnwidth}{!}{
{\tiny
\begin{tabular}{clllcccc}
\hline
\multicolumn{3}{c}{\multirow{2}{*}{\textbf{Methods}}}     &  & \multicolumn{3}{c}{\textbf{mAUC}{$\uparrow$}}       &  \\ \cline{5-7}
\multicolumn{3}{c}{}                              
&  & \textbf{HPatches} & \textbf{MegaDepth} & \textbf{ScanNet} &  \\ \hline
\multicolumn{2}{l}{LoFTR~\cite{sun2021loftr}}  
&  &  &    75.4    &   67.7         &     40.7      &    \\
\multicolumn{2}{l}{LightGlue~\cite{lindenberger2023lightglue}}  
&  &  &    \underline{77.5}    &   72.3         &     \underline{48.2}      &    \\
\multicolumn{2}{l}{RoMa~\cite{edstedt2024roma}}        
&  &  &    \textbf{78.4}    &  \textbf{75.2}          &     \textbf{52.0}      &    \\ 
\multicolumn{2}{l}{PDM (ours)}                
&  &  &    77.0    &  \underline{72.4}          &     46.5      &   \\
\hline
\end{tabular}
}}}
\scriptsize{
The bold and underline values denote the best and second best results, respectively.
}
\end{table}

\section{Discussion}
Despite the strong alignment performance of \textsc{PDM}, several limitations remain and warrant further investigation. 

First, the method is inherently reliant on keypoint detection, rendering it sensitive to challenging imaging conditions such as blur and low illumination. 
As illustrated in Fig.~\ref{fig:FIG_EXTR1} and Fig.~\ref{fig:FIG_EXTR2}, \textsc{PDM} demonstrates strong robustness in challenging scenarios including less-overlapping, rotated, or textureless regions by leveraging discriminative features from surrounding areas, effectively handling conditions commonly encountered in real-world applications.
Furthermore, \textsc{PDM} maintains stable correspondence estimation even when the query points on the source image are unevenly distributed or concentrated within a localized region, indicating resilience to spatial imbalance. 
As shown in Fig.~\ref{fig:FIG_FAIL}, \textsc{PDM} successfully estimates transformations in moderately difficult SFI-UWFI cases where GeoFormer~\cite{liu2023geometrized} fails. 
However, in more extreme scenarios where reliable keypoints cannot be detected, \textsc{PDM} also fails to achieve accurate alignment. 
Such cases are not uncommon in clinical settings, particularly when dealing with severely degraded UWFIs, highlighting the need to further enhance the framework’s robustness under poor imaging conditions.
Moreover, as demonstrated in Tab.~\ref{tab:ablative}, when local non-rigid deformation becomes excessively large and the number of valid correspondences is extremely limited (e.g., 4 or 50 particles), the performance drops notably, indicating that sparse correspondence density cannot fully capture fine-grained geometric variations, leading to partial misalignment.
Therefore, this reflects a limitation of PDM, which relies on a sufficiently dense particle sampling to compensate for the absence of explicit local deformation estimation.

Second, \textsc{PDM} may be susceptible to domain shifts between training and testing datasets, which can limit its  generalizability across diverse patient populations and imaging protocols. 
Moreover, the current implementation uses a fixed number of sampling iterations regardless of the complexity of the image pair, potentially resulting in inefficiencies such as over-computation for simpler cases or insufficient refinement for more complex alignment tasks.

Third, since this study aims to align SFIs to UWFIs, the query generator in \textsc{PDM} performs keypoint detection on the source image, the SFI.
SFIs generally possess rich texture, which facilitates effective keypoint detection.
However, when sufficient keypoints cannot be detected due to degradation or lack of information within the SFI, learnable feature extraction algorithms such as XFeat~\cite{potje2024cvpr} or SuperPoint~\cite{detone2018superpoint} could be considered.
Furthermore, to enhance global registration stability, future work could explore spatial regularization strategies that penalize particles drifting toward image boundaries or non-overlapping regions. Such region-aware constraints during training would encourage particles to maintain positions within valid correspondence regions, potentially improving convergence stability and registration accuracy.
In our \textsc{PDM}, since the output of the query generator serves as the condition for the diffusion model, it can be easily integrated with such learnable feature extraction models.

Finally, to examine the generalizability of our approach beyond medical imaging, we additionally evaluated \textsc{PDM} on three public scene correspondence benchmarks: HPatches~\cite{balntas2017hpatches}, MegaDepth~\cite{MegaDepthLi18}, and ScanNet~\cite{dai2017scannet}, as shown in Fig.~\ref{fig:FIG_HP}.
HPatches includes planar image pairs with variations in illumination and viewpoint, while MegaDepth and ScanNet contain outdoor and indoor scenes involving 3D epipolar transformations.
Following prior works\cite{zhao2021deep,sun2021loftr,edstedt2024roma}, we report the mean area under the curve (mAUC) of geometric accuracy for planar datasets, computed from the average corner error (ACE) at thresholds of 3, 5, and 10 pixels, and for two-view datasets, computed from the pose error at $5^\circ$, $10^\circ$, and $20^\circ$ based on rotation and translation deviations.
All models, including the baselines and our \textsc{PDM}, were trained on the Oxford–Paris datasets\cite{4270197,4587635} for fair comparison.
As shown in Tab.\ref{tab:scenery}, \textsc{PDM} obtains 77.0\%, 72.4\%, and 46.5\% mAUC on HPatches, MegaDepth, and ScanNet, respectively.
The results suggest that \textsc{PDM} maintains reasonable accuracy for planar transformations and exhibits a certain degree of transferability to 3D-structured scenes.
The relatively lower performance on the latter datasets may be attributed to the model’s training strategy with homography-based regularization and image pair containing depth and pose noise in monocular reconstruction, which does not fully capture complex 3D geometric relationships.
Overall, these experiments indicate that the proposed framework can be applied beyond specialized medical registration tasks, albeit with room for improvement in highly non-planar scenarios.

Future work will focus on enhancing robustness to severe image degradations, improving adaptability to domain shifts, and incorporating adaptive sampling strategies based on alignment complexity. These are critical steps toward reliable deployment in real-world medical settings.

\section{Conclusion}
We introduce \textsc{PDM}, a novel iterative random walk framework for retinal image alignment that unifies global-local alignment through a stochastic search process, progressively refining transformations by addressing both global misalignment and fine-grained local deformations.
\textsc{PDM} demonstrates robust and precise alignment capabilities, particularly excelling in the challenging task of aligning standard fundus images (SFIs) and ultra-widefield fundus images (UWFIs).
Our comparative evaluations show that \textsc{PDM} achieves state-of-the-art performance, surpassing recent methods in alignment accuracy.

\bibliographystyle{unsrtnat}
\bibliography{main}

@String(IJCV = {Int. J. Comput. Vis.})

@String(CVPR= {IEEE Conf. Comput. Vis. Pattern Recog.})

@String(ICCV= {Int. Conf. Comput. Vis.})

@String(ECCV= {Eur. Conf. Comput. Vis.})

@String(NeurIPS= {Adv. Neural Inform. Process. Syst.})

@String(ICLR = {Int. Conf. Learn. Represent.})

@String(CVPRW= {IEEE Conf. Comput. Vis. Pattern Recog. Worksh.})

@String(IJCV  = {IJCV})

@String(CVPR  = {CVPR})

@String(ICCV  = {ICCV})

@String(ECCV  = {ECCV})

@String(NeurIPS  = {NeurIPS})

@String(ICLR  = {ICLR})

@String(CVPRW= {CVPRW})

@article{lowe2004distinctive,
  title={Distinctive image features from scale-invariant keypoints},
  author={Lowe, David G},
  journal={IJCV},
  year={2004}
}

@inproceedings{bay2006surf,
  title={Surf: Speeded up robust features},
  author={Bay, Herbert and Tuytelaars, Tinne and Van Gool, Luc},
  booktitle={ECCV},
  year={2006}
}

@article{rosten2008faster,
  title={Faster and better: A machine learning approach to corner detection},
  author={Rosten, Edward and Porter, Reid and Drummond, Tom},
  journal={IEEE TPAMI},
  year={2008}
}

@article{he2015deepresiduallearningimage,
  title={Deep Residual Learning for Image Recognition}, 
  author={Kaiming He and Xiangyu Zhang and Shaoqing Ren and Jian Sun},
  year={2015},
  journal={arXiv}, 
}

@inproceedings{4723207,
  author={Zhang, Jun and Hu, Jinglu},
  booktitle={CSSE}, 
  title={Image Segmentation Based on 2D Otsu Method with Histogram Analysis}, 
  year={2008},
}

@inproceedings{10_1007,
    author={Farneb{\"a}ck, Gunnar},
    title={Two-Frame Motion Estimation Based on Polynomial Expansion},
    booktitle={Image Analysis},
    year={2003}
}

@inproceedings{calonder2010brief,
  title={Brief: Binary robust independent elementary features},
  author={Calonder, Michael and Lepetit, Vincent and Strecha, Christoph and Fua, Pascal},
  booktitle={ECCV},
  year={2010}
}

@inproceedings{jiang2021cotr,
  title={{COTR: Correspondence Transformer for Matching Across Images}},
  author={Wei Jiang and Eduard Trulls and Jan Hosang and Andrea Tagliasacchi and Kwang Moo Yi},
  booktitle={ICCV},
  year={2021}
}

@inproceedings{detone2018superpoint,
  title={Superpoint: Self-supervised interest point detection and description},
  author={DeTone, Daniel and Malisiewicz, Tomasz and Rabinovich, Andrew},
  booktitle={CVPRW},
  year={2018}
}

@misc{li2025comatchdynamiccovisibilityawaretransformer,
      title={CoMatch: Dynamic Covisibility-Aware Transformer for Bilateral Subpixel-Level Semi-Dense Image Matching}, 
      author={Zizhuo Li and Yifan Lu and Linfeng Tang and Shihua Zhang and Jiayi Ma},
      year={2025},
      eprint={2503.23925},
      archivePrefix={arXiv},
      primaryClass={cs.CV},
      url={https://arxiv.org/abs/2503.23925}, 
}

@INPROCEEDINGS{potje2024cvpr,
  author={Potje, Guilherme and Cadar, Felipe and Araujo, André and Martins, Renato and Nascimento, Erickson R.},
  booktitle={2024 IEEE/CVF Conference on Computer Vision and Pattern Recognition (CVPR)}, 
  title={XFeat: Accelerated Features for Lightweight Image Matching}, 
  year={2024},
  pages={2682-2691},
  doi={10.1109/CVPR52733.2024.00259}}

@ARTICLE{10643351,
  author={Li, Zizhuo and Zhang, Shihua and Ma, Jiayi},
  journal={IEEE Transactions on Pattern Analysis and Machine Intelligence}, 
  title={U-Match: Exploring Hierarchy-Aware Local Context for Two-View Correspondence Learning}, 
  year={2024},
  volume={46},
  number={12},
  pages={10960-10977},
  doi={10.1109/TPAMI.2024.3447048}}

@inproceedings{tuzcuouglu2024xoftr,
  title={XoFTR: Cross-modal Feature Matching Transformer},
  author={Tuzcuo{\u{g}}lu, {\"O}nder and K{\"o}ksal, Aybora and Sofu, Bu{\u{g}}ra and Kalkan, Sinan and Alatan, A Aydin},
  booktitle={Proceedings of the IEEE/CVF Conference on Computer Vision and Pattern Recognition},
  pages={4275--4286},
  year={2024}
}

@inproceedings{truong2019glampoints,
  title={Glampoints: Greedily learned accurate match points},
  author={Truong, Prune and Apostolopoulos, Stefanos and Mosinska, Agata and Stucky, Samuel and Ciller, Carlos and Zanet, Sandro De},
  booktitle={ICCV},
  year={2019}
}

@article{edstedt2024roma,
title={{RoMa: Robust Dense Feature Matching}},
author={Edstedt, Johan and Sun, Qiyu and Bökman, Georg and Wadenbäck, Mårten and Felsberg, Michael},
journal={IEEE Conference on Computer Vision and Pattern Recognition},
year={2024}
}

@inproceedings{ren2025minima,
  title={MINIMA: Modality Invariant Image Matching},
  author={Ren, Jiangwei and Jiang, Xingyu and Li, Zizhuo and Liang, Dingkang and Zhou, Xin and Bai, Xiang},
  booktitle={Proceedings of the IEEE/CVF Conference on Computer Vision and Pattern Recognition},
  year={2025}
}

@inProceedings{MegaDepthLi18,
  title={MegaDepth: Learning Single-View Depth Prediction from Internet Photos},
  author={Zhengqi Li and Noah Snavely},
  booktitle={Computer Vision and Pattern Recognition (CVPR)},
  year={2018}
}

@INPROCEEDINGS{4270197,
  author={Philbin, James and Chum, Ondrej and Isard, Michael and Sivic, Josef and Zisserman, Andrew},
  booktitle={2007 IEEE Conference on Computer Vision and Pattern Recognition}, 
  title={Object retrieval with large vocabularies and fast spatial matching}, 
  year={2007},
  volume={},
  number={},
  pages={1-8},
  doi={10.1109/CVPR.2007.383172}}

@inproceedings{dai2017scannet,
    title={ScanNet: Richly-annotated 3D Reconstructions of Indoor Scenes},
    author={Dai, Angela and Chang, Angel X. and Savva, Manolis and Halber, Maciej and Funkhouser, Thomas and Nie{\ss}ner, Matthias},
    booktitle = {Proc. Computer Vision and Pattern Recognition (CVPR), IEEE},
    year = {2017}

}

@INPROCEEDINGS{4587635,
  title={Lost in quantization: Improving particular object retrieval in large scale image databases},
  author={Philbin, James and Chum, Ondrej and Isard, Michael and Sivic, Josef and Zisserman, Andrew},
  booktitle={2008 IEEE Conference on Computer Vision and Pattern Recognition},
  year={2008},
  pages={1--8},
  doi={10.1109/CVPR.2008.4587635}
}

@article{dhariwal2021diffusionmodelsbeatgans,
  title={Diffusion Models Beat GANs on Image Synthesis}, 
  author={Prafulla Dhariwal and Alex Nichol},
  year={2021},
  journal={arXiv} 
}

@article{revaud2019r2d2,
  title={R2d2: Reliable and repeatable detector and descriptor},
  author={Revaud, Jerome and De Souza, Cesar and Humenberger, Martin and Weinzaepfel, Philippe},
  journal={NeurIPS},
  year={2019}
}

@inproceedings{sarlin2020superglue,
  title={Superglue: Learning feature matching with graph neural networks},
  author={Sarlin, Paul-Edouard and DeTone, Daniel and Malisiewicz, Tomasz and Rabinovich, Andrew},
  booktitle={CVPR},
  year={2020}
}

@inproceedings{liu2022semi,
  title={Semi-supervised Keypoint Detector and Descriptor for Retinal Image Matching},
  author={Liu, Jiazhen and Li, Xirong and Wei, Qijie and Xu, Jie and Ding, Dayong},
  booktitle={ECCV},
  year={2022}
}

@inproceedings{lindenberger2023lightglue,
  author    = {Philipp Lindenberger and Paul-Edouard Sarlin and Marc Pollefeys},
  title     = {{LightGlue: Local Feature Matching at Light Speed}},
  booktitle = {ICCV},
  year      = {2023}
}

@article{detone2016deep,
  title={Deep image homography estimation},
  author={DeTone, Daniel and Malisiewicz, Tomasz and Rabinovich, Andrew},
  journal={arXiv},
  year={2016}
}

@inproceedings{sun2021loftr,
  title={LoFTR: Detector-free local feature matching with transformers},
  author={Sun, Jiaming and Shen, Zehong and Wang, Yuang and Bao, Hujun and Zhou, Xiaowei},
  booktitle={CVPR},
  year={2021}
}

@inproceedings{dosovitskiy2015flownet,
  title={Flownet: Learning optical flow with convolutional networks},
  author={Dosovitskiy, Alexey and Fischer, Philipp and Ilg, Eddy and Hausser, Philip and Hazirbas, Caner and Golkov, Vladimir and Van Der Smagt, Patrick and Cremers, Daniel and Brox, Thomas},
  booktitle={ICCV},
  year={2015}
}

@inproceedings{weinzaepfel2013deepflow,
  title={DeepFlow: Large displacement optical flow with deep matching},
  author={Weinzaepfel, Philippe and Revaud, Jerome and Harchaoui, Zaid and Schmid, Cordelia},
  booktitle={ICCV},
  year={2013}
}

@article{jaderberg2015spatial,
  title={Spatial transformer networks},
  author={Jaderberg, Max and Simonyan, Karen and Zisserman, Andrew and others},
  journal={NeurIPS},
  year={2015}
}

@inproceedings{noh2019fine,
  title={Fine-scale vessel extraction in fundus images by registration with fluorescein angiography},
  author={Noh, Kyoung Jin and Park, Sang Jun and Lee, Soochahn},
  booktitle={MICCAI},
  year={2019}
}

@article{de2019deep,
  title={A deep learning framework for unsupervised affine and deformable image registration},
  author={De Vos, Bob D and Berendsen, Floris F and Viergever, Max A and Sokooti, Hessam and Staring, Marius and I{\v{s}}gum, Ivana},
  journal={MedIA},
  year={2019}
}

@article{lee2016ultra,
  title={Ultra-widefield retina imaging: principles of technology and clinical applications},
  author={Lee, Junyeop and Sagong, Min},
  journal={Journal of Retina},
  year={2016}
}

@InProceedings{lee2019adeep,
author = {Lee, Jimmy Addison and Liu, Peng and Cheng, Jun and Fu, Huazhu},
title = {A Deep Step Pattern Representation for Multimodal Retinal Image Registration},
booktitle = {ICCV},
year = {2019}
}

@article{witmer2013comparison,
  title={Comparison of ultra-widefield fluorescein angiography with the Heidelberg Spectralis{\textregistered} noncontact ultra-widefield module versus the Optos{\textregistered} Optomap{\textregistered}},
  author={Witmer, Matthew T and Parlitsis, George and Patel, Sarju and Kiss, Szil{\'a}rd},
  journal={Clinical Ophthalmology},
  year={2013}
}

@article{dhariwal2021diffusion,
  title={Diffusion models beat gans on image synthesis},
  author={Dhariwal, Prafulla and Nichol, Alexander},
  journal={NeurIPS},
  year={2021}
}

@inproceedings{kim2022diffusemorph,
  title={DiffuseMorph: unsupervised deformable image registration using diffusion model},
  author={Kim, Boah and Han, Inhwa and Ye, Jong Chul},
  booktitle={ECCV},
  year={2022}
}

@inproceedings{Brachmann2019DSACstar,
  title     = {DSAC* - Differentiable RANSAC for Camera Localization},
  author    = {Brachmann et al.},
  booktitle = {CVPR},
  year      = {2019},

}

@inproceedings{rombach2022high,
  title={High-resolution image synthesis with latent diffusion models},
  author={Rombach, Robin and Blattmann, Andreas and Lorenz, Dominik and Esser, Patrick and Ommer, Bj{\"o}rn},
  booktitle={CVPR},
  year={2022}
}

@article{hernandez2020rempe,
  title={REMPE: Registration of retinal images through eye modelling and pose estimation},
  author={Hernandez-Matas, Carlos and Zabulis, Xenophon and Argyros, Antonis A},
  journal={IEEE JBHI},
  year={2020}
}

@article{kim2021cyclemorph,
  title={CycleMorph: cycle consistent unsupervised deformable image registration},
  author={Kim, Boah and Kim, Dong Hwan and Park, Seong Ho and Kim, Jieun and Lee, June-Goo and Ye, Jong Chul},
  journal={MedIA},
  year={2021},

}

@inproceedings{zhao2025diffusionsfm,
  title={DiffusionSfM: Predicting Structure and Motion via Ray Origin and Endpoint Diffusion},
  author={Zhao, Qitao and Lin, Amy and Tan, Jeff and Zhang, Jason Y and Ramanan, Deva and Tulsiani, Shubham},
  booktitle={CVPR},
  pages={6317--6326},
  year={2025}
}

@inproceedings{xu2019deepatlas,
  title={DeepAtlas: Joint semi-supervised learning of image registration and segmentation},
  author={Xu, Zhenlin and Niethammer, Marc},
  booktitle={MICCAI},
  year={2019}
}

@article{hu2018weakly,
  title={Weakly-supervised convolutional neural networks for multimodal image registration},
  author={Hu, Yipeng and Modat, Marc and Gibson, Eli and Li, Wenqi and Ghavami, Nooshin and Bonmati, Ester and Wang, Guotai and Bandula, Steven and Moore, Caroline M and Emberton, Mark and others},
  journal={MedIA},
  year={2018}
}

@inproceedings{cao2017deformable,
  title={Deformable image registration based on similarity-steered CNN regression},
  author={Cao, Xiaohuan and Yang, Jianhua and Zhang, Jun and Nie, Dong and Kim, Minjeong and Wang, Qian and Shen, Dinggang},
  booktitle={MICCAI},
  year={2017}
}

@inproceedings{xu2022gmflow,
  title={Gmflow: Learning optical flow via global matching},
  author={Xu, Haofei and Zhang, Jing and Cai, Jianfei and Rezatofighi, Hamid and Tao, Dacheng},
  booktitle={CVPR},
  year={2022}
}

@inproceedings{zhang2021separable,
  title={Separable flow: Learning motion cost volumes for optical flow estimation},
  author={Zhang, Feihu and Woodford, Oliver J and Prisacariu, Victor Adrian and Torr, Philip HS},
  booktitle={ICCV},
  year={2021}
}

@inproceedings{huang2022flowformer,
  title={Flowformer: A transformer architecture for optical flow},
  author={Huang, Zhaoyang and Shi, Xiaoyu and Zhang, Chao and Wang, Qiang and Cheung, Ka Chun and Qin, Hongwei and Dai, Jifeng and Li, Hongsheng},
  booktitle={ECCV},
  year={2022}
}

@inproceedings{liu2023geometrized,
  title={Geometrized Transformer for Self-Supervised Homography Estimation},
  author={Liu, Jiazhen and Li, Xirong},
  booktitle={ICCV},
  year={2023}
}

@article{wu2024diff,
  title={Diff-Reg v1: Diffusion Matching Model for Registration Problem},
  author={Wu, Qianliang and Jiang, Haobo and Luo, Lei and Li, Jun and Ding, Yaqing and Xie, Jin and Yang, Jian},
  journal={arXiv preprint arXiv:2403.19919},
  year={2024}
}

@inproceedings{wang2023posediffusion,
  title={PoseDiffusion: Solving Pose Estimation via Diffusion-aided Bundle Adjustment},
  author={Wang, Jianyuan and Rupprecht, Christian and Novotny, David},
  booktitle={ICCV},
  year={2023}
}

@misc{Li2021Flori,
    author = {Ding, Li and Kang, Tony and Kuriyan, Ajay and Ramchandran, Rajeev and Wykoff, Charles and Sharma, Gaurav},
    title = {FLoRI21: Fluorescein Angiography Longitudinal Retinal Image Registration Dataset},
    journal={IEEE Dataport},
    year = {2021} 
}

@misc{sivaraman2024retinaregnetzeroshotapproachretinal,
      title={RetinaRegNet: A Zero-Shot Approach for Retinal Image Registration}, 
      author={Vishal Balaji Sivaraman and Muhammad Imran and Qingyue Wei and Preethika Muralidharan and Michelle R. Tamplin and Isabella M . Grumbach and Randy H. Kardon and Jui-Kai Wang and Yuyin Zhou and Wei Shao},
      year={2024},
      journal={arXiv} 
}

@article{rocco2020ncnet,
  title={Ncnet: Neighbourhood consensus networks for estimating image correspondences},
  author={Rocco, Ignacio and Cimpoi, Mircea and Arandjelovi{\'c}, Relja and Torii, Akihiko and Pajdla, Tomas and Sivic, Josef},
  journal={IEEE TPAMI},
  year={2020}
}

@article{ho2020denoising,
  title={Denoising diffusion probabilistic models},
  author={Ho, Jonathan and Jain, Ajay and Abbeel, Pieter},
  journal={NeurIPS},
  year={2020}
}

@inproceedings{zhang2022relpose,
  title={Relpose: Predicting probabilistic relative rotation for single objects in the wild},
  author={Zhang, Jason Y and Ramanan, Deva and Tulsiani, Shubham},
  booktitle={ECCV},
  year={2022}
}

@inproceedings{sinha2023sparsepose,
  title={Sparsepose: Sparse-view camera pose regression and refinement},
  author={Sinha, Samarth and Zhang, Jason Y and Tagliasacchi, Andrea and Gilitschenski, Igor and Lindell, David B},
  booktitle={CVPR},
  year={2023}
}

@article{hernandezmatas2017fire, 
  title={FIRE: Fundus Image Registration dataset}, 
  journal={Modeling and Artificial Intelligence in Ophthalmology},
  author={Hernandez-Matas, Carlos and Zabulis, Xenophon and Triantafyllou, Areti and Anyfanti, Panagiota and Douma, Stella and Argyros, Antonis A}, 
  year={2017},
}

@inproceedings{chen2022aspanformer,
  author    = {Chen, Hongkai and Luo, Zixin and Zhou, Lei and Tian, Yurun and Zhen, Mingmin and Fang, Tian and McKinnon, David and Tsin, Yanghai and Quan, Long},
  title     = {ASpanFormer: Detector-Free Image Matching with Adaptive Span Transformer},
  booktitle   = {ECCV},
  year      = {2022},
}

@inproceedings{
  song2021scorebased,
  title={Score-Based Generative Modeling through Stochastic Differential Equations},
  author={Yang Song and Jascha Sohl-Dickstein and Diederik P Kingma and Abhishek Kumar and Stefano Ermon and Ben Poole},
  booktitle={ICLR},
  year={2021}
}

@misc{sohldickstein2015deep,
  title={Deep Unsupervised Learning using Nonequilibrium Thermodynamics}, 
  author={Jascha Sohl-Dickstein and Eric A. Weiss and Niru Maheswaranathan and Surya Ganguli},
  year={2015},
  journal={arXiv} 
}

@inproceedings{lee2019istn,
    author = {Lee, Matthew C.H. and Oktay, Ozan and Schuh, Andreas and Schaap, Michiel and Glocker, Ben},
    title = {Image-and-Spatial Transformer Networks for Structure-guided Image Registration},
    year = {2019},
    booktitle = {MICCAI}
}

@article{BahadarKhan2016morph,
    author = {BahadarKhan, Khan AND A Khaliq, Amir AND Shahid, Muhammad},
    journal = {Plos one},
    publisher = {Public Library of Science},
    title = {A Morphological Hessian Based Approach for Retinal Blood Vessels Segmentation and Denoising Using Region Based Otsu Thresholding},
    year = {2016}

}

@article{cootes1995active,
  title={Active shape models-their training and application},
  author={Cootes, Timothy F and Taylor, Christopher J and Cooper, David H and Graham, Jim},
  journal={Computer vision and image understanding},
  year={1995},
}

@article{chan2024tutorial,
  title={Tutorial on Diffusion Models for Imaging and Vision},
  author={Chan, Stanley H},
  journal={arXiv},
  year={2024}
}

@inproceedings{besl1992method,
  title={Method for registration of 3-D shapes},
  author={Besl, Paul J and McKay, Neil D},
  booktitle={Sensor fusion IV: control paradigms and data structures},
  year={1992},
}

@article{fischler1981random,
  title={Random sample consensus: a paradigm for model fitting with applications to image analysis and automated cartography},
  author={Fischler, Martin A and Bolles, Robert C},
  journal={Communications of the ACM},
  year={1981},
}

@InProceedings{zhang2024raydiffusion,
    title={Cameras as Rays: Pose Estimation via Ray Diffusion},
    author={Zhang, Jason Y and Lin, Amy and Kumar, Moneish and Yang, Tzu-Hsuan and Ramanan, Deva and Tulsiani, Shubham},
    booktitle={ICLR},
    year={2024}
}

@inproceedings{dong2018learning,
  title={Learning to align images using weak geometric supervision},
  author={Dong, Jing and Boots, Byron and Dellaert, Frank and Chandra, Ranveer and Sinha, Sudipta},
  booktitle={3DV},
  year={2018},
}

@inproceedings{cao2022iterative,
  title={Iterative deep homography estimation},
  author={Cao, Si-Yuan and Hu, Jianxin and Sheng, Zehua and Shen, Hui-Liang},
  booktitle={CVPR},
  year={2022}
}

@inproceedings{cao2023recurrent,
  title={Recurrent homography estimation using homography-guided image warping and focus transformer},
  author={Cao, Si-Yuan and Zhang, Runmin and Luo, Lun and Yu, Beinan and Sheng, Zehua and Li, Junwei and Shen, Hui-Liang},
  booktitle={CVPR},
  year={2023}
}

@inproceedings{zhu2024mcnet,
  title={MCNet: Rethinking the Core Ingredients for Accurate and Efficient Homography Estimation},
  author={Zhu, Haokai and Cao, Si-Yuan and Hu, Jianxin and Zuo, Sitong and Yu, Beinan and Ying, Jiacheng and Li, Junwei and Shen, Hui-Liang},
  booktitle={CVPR},
  year={2024}
}

@misc{balntas2017hpatches,
      title={HPatches: A benchmark and evaluation of handcrafted and learned local descriptors}, 
      author={Vassileios Balntas and Karel Lenc and Andrea Vedaldi and Krystian Mikolajczyk},
      year={2017},
      journal={arXiv} 
}

@inproceedings{zhao2021deep,
  title={Deep Lucas-Kanade Homography for Multimodal Image Alignment},
  author={Zhao, Yiming and Huang, Xinming and Zhang, Ziming},
  booktitle={CVPR},
  year={2021}
}

@ARTICLE{deng2024crosshomo,
  author={Deng, Xin and Liu, Enpeng and Gao, Chao and Li, Shengxi and Gu, Shuhang and Xu, Mai},
  journal={IEEE Transactions on Pattern Analysis and Machine Intelligence}, 
  title={CrossHomo: Cross-Modality and Cross-Resolution Homography Estimation}, 
  year={2024}
}

@article{lee2023deep,
  title={A deep learning-based framework for retinal fundus image enhancement},
  author={Lee, Kang Geon and Song, Su Jeong and Lee, Soochahn and Yu, Hyeong Gon and Kim, Dong Ik and Lee, Kyoung Mu},
  journal={Plos one},
  year={2023}
}

@article{ju2020bridge,
  title={Bridge the domain gap between ultra-wide-field and traditional fundus images via adversarial domain adaptation},
  author={Ju, Lie and Wang, Xin and Zhou, Quan and Zhu, Hu and Harandi, Mehrtash and Bonnington, Paul and Drummond, Tom and Ge, Zongyuan},
  journal={arXiv},
  year={2020}
}

@article{yoo2020deep,
  title={Deep learning can generate traditional retinal fundus photographs using ultra-widefield images via generative adversarial networks},
  author={Yoo, Tae Keun and Ryu, Ik Hee and Kim, Jin Kuk and Lee, In Sik and Kim, Jung Sub and Kim, Hong Kyu and Choi, Joon Yul},
  journal={CMPB},
  year={2020},
}

@article{thuma2023big,
  title={The big warp: Registration of disparate retinal imaging modalities and an example overlay of ultrawide-field photos and en-face OCTA images},
  author={Thuma, Tobin BT and Bogovic, John A and Gunton, Kammi B and Jimenez, Hiram and Negreiros, Bernardo and Pulido, Jose S},
  journal={Plos one},
  year={2023},
}

@inproceedings{lim2017enhanced,
  title={Enhanced deep residual networks for single image super-resolution},
  author={Lim, Bee and Son, Sanghyun and Kim, Heewon and Nah, Seungjun and Mu Lee, Kyoung},
  booktitle={CVPRW},
  year={2017}
}

@article{balakrishnan2019voxelmorph,
  title={Voxelmorph: a learning framework for deformable medical image registration},
  author={Balakrishnan, Guha and Zhao, Amy and Sabuncu, Mert R and Guttag, John and Dalca, Adrian V},
  journal={IEEE TMI},
  year={2019},
}

@book{inbookHaynes,
author = {Haynes, Winston},
year = {2013},
month = {01},
pages = {2354-2355},
title = {Wilcoxon Rank Sum Test},
isbn = {978-1-4419-9862-0},
doi = {10.1007/978-1-4419-9863-7_1185}
}

@ARTICLE{10158729,
  author={Chen, Zeyuan and Zheng, Yuanjie and Gee, James C.},
  journal={IEEE Transactions on Medical Imaging}, 
  title={TransMatch: A Transformer-Based Multilevel Dual-Stream Feature Matching Network for Unsupervised Deformable Image Registration}, 
  year={2024},
  volume={43},
  number={1},
  pages={15-27},
  doi={10.1109/TMI.2023.3288136}}

@INPROCEEDINGS {10821896,
author = { Liu, Yepeng and Yu, Baosheng and Chen, Tian and Gu, Yuliang and Du, Bo and Xu, Yongchao and Cheng, Jun },
booktitle = { 2024 IEEE International Conference on Bioinformatics and Biomedicine (BIBM) },
title = {{ Progressive Retinal Image Registration via Global and Local Deformable Transformations }},
year = {2024},
volume = {},
ISSN = {},
pages = {2183-2190},
doi = {10.1109/BIBM62325.2024.10821896}
}

@Article{bioengineering11060568,
AUTHOR = {Lee, Kang Geon and Song, Su Jeong and Lee, Soochahn and Kim, Bo Hee and Kong, Mingui and Lee, Kyoung Mu},
TITLE = {FQ-UWF: Unpaired Generative Image Enhancement for Fundus Quality Ultra-Widefield Retinal Images},
JOURNAL = {Bioengineering},
VOLUME = {11},
YEAR = {2024},
ARTICLE-NUMBER = {568}
}

@article{ZhangWB15,
  author       = {Min Zhang and
                  Teresa Wu and
                  Kevin M. Bennett},
  title        = {Small Blob Identification in Medical Images Using Regional Features
                  From Optimum Scale},
  journal      = {IEEE TBME},
  year         = {2015},
  
}

@inproceedings{Harris88alvey,
  author    = {Christopher G. Harris and Mike Stephens},
  title     = {A Combined Corner and Edge Detector},
  booktitle = {Proceedings of the 4th Alvey Vision Conference},
  pages     = {147--151},
  year      = {1988},
  publisher = {Alvey Vision Club},
  doi       = {10.5244/C.2.23},
}

@inproceedings{RostenDrummond2006,
  author    = {Edward Rosten and Tom Drummond},
  title     = {Machine learning for high‑speed corner detection},
  booktitle = {ECCV Workshop on Advanced Video-Based Surveillance},
  year      = {2006},
}

@inproceedings{Alcantarilla2012KAZE,
  author    = {Pablo F. Alcantarilla and Adrien Bartoli and Andrew J. Davison},
  title     = {KAZE Features},
  booktitle = {European Conference on Computer Vision (ECCV)},
  pages     = {214--227},
  year      = {2012},
  publisher = {Springer},
}

@misc{aithal2024understandinghallucinationsdiffusionmodels,
      title={Understanding Hallucinations in Diffusion Models through Mode Interpolation}, 
      author={Sumukh K Aithal and Pratyush Maini and Zachary C. Lipton and J. Zico Kolter},
      year={2024},
      eprint={2406.09358},
      archivePrefix={arXiv},
      primaryClass={cs.LG}, 
}

\end{document}